\documentclass{article}

\usepackage{microtype}
\usepackage{graphicx}
\usepackage{subcaption}
\usepackage{booktabs} % for professional tables

\usepackage{hyperref}

\usepackage[preprint]{icml2026}

\usepackage{amsmath}
\usepackage{amssymb}
\usepackage{mathtools}
\usepackage{amsthm}

\usepackage[capitalize,noabbrev]{cleveref}

\theoremstyle{plain}

\theoremstyle{definition}

\theoremstyle{remark}

\usepackage[textsize=tiny]{todonotes}

\icmltitlerunning{GPT2MEG: Quantizing MEG for Autoregressive Generation}

\begin{document}

\twocolumn[
  \icmltitle{GPT2MEG: Quantizing MEG for Autoregressive Generation}

  % It is OKAY to include author information, even for blind submissions: the
  % style file will automatically remove it for you unless you've provided
  % the [accepted] option to the icml2026 package.

  % List of affiliations: The first argument should be a (short) identifier you
  % will use later to specify author affiliations Academic affiliations
  % should list Department, University, City, Region, Country Industry
  % affiliations should list Company, City, Region, Country

  % You can specify symbols, otherwise they are numbered in order. Ideally, you
  % should not use this facility. Affiliations will be numbered in order of
  % appearance and this is the preferred way.
  \icmlsetsymbol{equal}{*}

  \begin{icmlauthorlist}
    \icmlauthor{Richard Csaky}{fore}
    \icmlauthor{Mats W.J. van Es}{ohba,wel}
    \icmlauthor{Oiwi Parker Jones}{wel,eng}
    \icmlauthor{Mark Woolrich}{ohba,wel}
  \end{icmlauthorlist}

  \icmlaffiliation{ohba}{Oxford Centre for Human Brain Activity, Department of Psychiatry, University of Oxford, Oxford, UK}
  \icmlaffiliation{wel}{Wellcome Centre for Integrative Neuroimaging, Oxford, UK}
  \icmlaffiliation{eng}{Department of Engineering Science, University of Oxford, Oxford, UK}
  \icmlaffiliation{fore}{Foresight Institute}

  \icmlcorrespondingauthor{Richard Csaky}{richard.csaky@gmail.com}

  % You may provide any keywords that you find helpful for describing your
  % paper; these are used to populate the "keywords" metadata in the PDF but
  % will not be shown in the document
  \icmlkeywords{Magnetoencephalography, Time series, Autoregressive modeling, Generative modeling, GPT, Tokenization, Conditional generation, multi-subject modeling}

  \vskip 0.3in
]

% this must go after the closing bracket ] following \twocolumn[ ...

% This command actually creates the footnote in the first column listing the
% affiliations and the copyright notice. The command takes one argument, which
% is text to display at the start of the footnote. The \icmlEqualContribution
% command is standard text for equal contribution. Remove it (just {}) if you
% do not need this facility.

% Use ONE of the following lines. DO NOT remove the command.
% If you have no special notice, KEEP empty braces:
\printAffiliationsAndNotice{}  % no special notice (required even if empty)
% Or, if applicable, use the standard equal contribution text:
% \printAffiliationsAndNotice{\icmlEqualContribution}

\begin{abstract}
Foundation models trained with self-supervised objectives are increasingly applied to brain recordings, but autoregressive generation of realistic multichannel neural time series remains comparatively underexplored, particularly for Magnetoencephalography (MEG). We study (i) modified multichannel WaveNet variants and (ii) a GPT-2-style Transformer, autoregressively trained by next-step prediction on unlabelled MEG. For the Transformer, we propose a simple quantization/tokenization and embedding scheme (channel, subject, and task-condition embeddings) that repurposes a language-model architecture for continuous, high-rate multichannel time series and enables conditional simulation of task-evoked activity. Across forecasting, long-horizon generation, and downstream decoding, \texttt{GPT2MEG} more faithfully reproduces temporal, spectral, and task-evoked statistics of real MEG than WaveNet variants and linear autoregressive baselines, and scales to multiple subjects via subject embeddings. %Code will be released on GitHub.
Code available at \url{https://github.com/ricsinaruto/MEG-transfer-decoding}.
\end{abstract}

\section{Introduction}

Magnetoencephalography (MEG) and related electrophysiology (EEG/ECoG) provide millisecond-resolved measurements of large-scale brain dynamics, but remain challenging to model due to high dimensionality, strong temporal structure, and substantial variability across subjects and recording setups. Recent self-supervised \emph{foundation models} have begun to learn transferable representations from large collections of unlabelled neural recordings, enabling improved performance and data efficiency on downstream decoding/encoding tasks \citep{kostas2021bendr, wang2023brainbert, cui2023neuro, jiang2024labram, wang2024eegpt, elouahidi2025reve}. Most existing brain foundation models focus on \emph{representation learning} via masked or contrastive objectives, producing embeddings for classification and regression. In contrast, \emph{autoregressive generative} foundation models, capable of simulating realistic neural time series and supporting data augmentation, uncertainty-aware decoding, and mechanistic hypothesis testing, have received comparatively less attention, especially for MEG.

Autoregressive sequence models are a natural fit for neural time series. Going back to the pre-foundation-model era, in audio, WaveNet modeled waveforms via causal dilated convolutions over discretized samples \citep{oord2016wavenet}. In language, decoder-only Transformers learn long-range dependencies through self-attention over discrete tokens \citep{Vaswani:2017, radford2019language}. Inspired by recent work on tokenized time-series models \citep{ansari2024chronos, das2024decoderonly, rasul2023lagllama}, we ask: \emph{can we repurpose autoregressive language-model training to build a generative "foundation"\footnote{While our experiments are small-scale, the framework has the potential to scale to produce foundational models.} model for multichannel MEG?}

We propose two classes of autoregressive models trained purely by next-step prediction on unlabelled MEG: (i) modified multichannel WaveNet variants and (ii) a modified GPT-2 model we call \texttt{GPT2MEG}. Our key contribution is a simple quantization/tokenization and embedding scheme that allows a discrete-token Transformer to model continuous multichannel MEG. We further incorporate time-aligned task-condition labels as local inputs, enabling conditional generation of task-evoked responses, and we scale to multiple subjects via subject embeddings \citep{csaky2023group}.

\textbf{Contributions.}
\begin{itemize}\itemsep0.25em
    \item We introduce \texttt{GPT2MEG}, an autoregressive Transformer model for sensor-level MEG using discrete tokenization with channel-, subject-, and stimulus-embeddings for conditional generation.
    \item We benchmark Transformer- and WaveNet-based generative models against a linear AR baseline, showing that one-step forecasting accuracy is insufficient; instead we evaluate long-horizon simulations via spectral metrics, HMM-derived dynamics, and task-evoked responses.
    \item We demonstrate multi-subject scaling with subject embeddings and show that simulated trials from a multi-subject model can improve downstream decoding through transfer learning.
\end{itemize}

\section{Related Work}

\textbf{Self-supervised and foundation models for electrophysiology.}
Self-supervised learning has enabled representation learning from unlabelled EEG/MEG using contrastive, masked, and predictive objectives \citep{banville2021uncovering, kostas2021bendr, wang2023brainbert}. Recent work scales these ideas towards \emph{foundation} models for brain signals \citep{cui2023neuro, jiang2024labram, wang2024eegpt, elouahidi2025reve}. Transformers are increasingly used in EEG analysis; for a recent review see \citet{vafaei2025transformers}. For MEG, related sequence models have also been explored for stimulus-to-brain encoding \citep{chehab2021deep}. These approaches primarily target embedding learning for decoding/encoding; our focus is complementary: building \emph{autoregressive generative} models that can simulate realistic multichannel MEG and task-evoked activity.

\textbf{Generative models for brain signals.}
Generative modelling of electrophysiology has been explored with VAEs, GANs, diffusion models, and Transformers, often motivated by data augmentation in BCI settings. Recent examples include transformer-based EEG synthesis with learned discrete codes \citep{lim2024eegtrans}. We study sensor-level MEG with explicit channel, subject, and task-condition embeddings, and we evaluate generation using multivariate dynamical metrics (HMM state statistics) in addition to spectral and evoked-response analyses.

\textbf{Tokenized time-series language models.}
Several recent foundation models for generic time-series forecasting tokenize continuous values into discrete vocabularies and train language-model architectures with cross-entropy loss, enabling zero-shot or few-shot generalization \citep{ansari2024chronos, das2024decoderonly, rasul2023lagllama}. Our \texttt{GPT2MEG} adopts a similarly simple discretization ($\mu$-law companding + uniform bins) but addresses challenges specific to MEG: high-dimensional sensor arrays, subject variability, and time-aligned task conditioning.

\section{Methods}
\label{sec:methods}

\subsection{Problem setup and notation}
\label{ssec:problem_setup}

We consider multichannel sensor-level MEG recordings $\mathbf{X}\in\mathbb{R}^{C\times T}$ with $C$ sensors and $T$ time points. After preprocessing and scaling, each sample is discretized into one of $Q$ tokens via $\mu$-law companding followed by uniform quantization (Section~\ref{ssec:full_wavenet}). For task datasets we also have a time-aligned condition sequence $\mathbf{y}_{1:T}$ (e.g.\ stimulus identity), and for multi-subject training a subject index $s$. Our goal is to learn a causal conditional generative model
\begin{equation}
p(\mathbf{X}\mid \mathbf{y}, s) = \prod_{t=1}^{T} p(\mathbf{x}_t \mid \mathbf{x}_{1:t-1}, \mathbf{y}_{\le t}, s),
\end{equation}
where $\mathbf{x}_t\in\mathbb{R}^{C}$ is the multichannel sample at time $t$. Some of the models we evaluate factorize across sensors (treating the channel dimension as a batch dimension during training), while others include explicit mixing across channels. All deep learning models are trained by teacher-forced next-token prediction with cross-entropy loss. We evaluate both one-step forecasting and long-horizon recursive generation using spectral, multivariate-dynamical, and task-evoked metrics (Section~\ref{ssec:gen_eval}).

\subsection{Multi-channel Wavenet}
\label{ssec:full_wavenet}

We adapt WaveNet \citep{oord2016wavenet} as an autoregressive baseline for multichannel MEG. The amplitude of each channel (and timestep) is discretized independently into $Q{=}256$ bins using $\mu$-law companding followed by uniform quantization. Concretely, after scaling each channel to $(-1,1)$ we apply

\begin{equation}
    f(x) = \mathrm{sign}(x)\,\frac{\ln(1 + \mu |x|)}{\ln(1 + \mu)},\quad \mu{=}255,
\end{equation}

and then uniformly bin $f(x)$ into $Q$ discrete tokens. Training with cross-entropy enables sampling from a learned categorical distribution and helps avoid mean-prediction bias from MSE regression \citep{banville2021uncovering}.

We evaluate two variants. \texttt{WavenetFullChannel} (\texttt{WFC}) treats channels as a batch dimension: one shared causal dilated-convolution stack is applied independently to each channel, with channel identity provided via a per-channel embedding of the discrete tokens. \texttt{WavenetFullChannelMix} (\texttt{WFCM}) additionally applies a learned $C\times C$ mixing projection to the skip representation to share information across channels. For task data we optionally provide time-aligned condition labels (and subject IDs for group training) through standard WaveNet local conditioning embeddings. Full equations are in Appendix~\ref{ssec:appendix_wavenet_old}.

\subsection{Multi-channel GPT2}
\label{ssec:transformers}

We set out to design a Transformer model suited for M/EEG data, while keeping the key elements that made it successful in language modelling. Specifically, we use GPT-2. When adapting it to continuous multivariate time series, the main challenges are at the input and output layers interfacing the model with the data.

To apply GPT2 to our continuous multichannel time series data, we take a similar approach as with Wavenet by tokenising each channel independently using the same method as before. This serves as our equivalent of the discrete set of tokens in language modelling. The same GPT2 model is applied to each channel in parallel by setting the channel dimension as the batch dimension. We call this \texttt{GPT2MEG}.

The input to the model includes the position embedding as well as subject and task-stimulus embeddings. We also add a label/embedding telling GPT2 which channel the current time series is coming from:

\begin{align}
    \mathbf{H}^{(0)} &= \mathbf{X}\mathbf{W}_e + \mathbf{W}_{p} + \mathbf{Y}\mathbf{W}_y + \mathbf{O}\mathbf{W}_o + \mathbf{W}_{c}
\end{align}

where $+$ denotes element-wise addition, $\mathbf{X} \in \mathbb{R}^{C \times T \times Q}$ is the tokenised input, $\mathbf{W}_{c} \in \mathbb{R}^{C \times E}$ are the learned channel embeddings of size $E$, which are distinct for each channel $c \in {1, \dots, C}$ but constant across time $t$. $\mathbf{Y}$ and $\mathbf{O}$ are the task and subject index matrices, mapped to their respective embeddings with $\mathbf{W}_{y} \in \mathbb{R}^{Y \times E}$ and $\mathbf{W}_{o} \in \mathbb{R}^{O \times E}$. $Y$ is the number of conditions and $O$ is the number of subjects. During timesteps when there is no stimulus present, $\mathbf{Y}\mathbf{W}_y$ is set to zero. As with the positional encoding $\mathbf{W}_p$, we simply add all embeddings (task, subject, channel) into a single representation. Note that instead of having channel-specific embeddings of the tokenised input $\mathbf{X}$ we learn the same mapping $\mathbf{W}_e \in \mathbb{R}^{Q \times E}$ across channels. Channel information is provided to the model through the channel embeddings.

A serious limitation of this channel-independent GPT2 model is that when predicting a single channel, it does not receive information from other channels. This is analogous to a univariate autoregressive model and ignores crucial cross-channel dependencies in the data. To be clear we often use the term univariate AR modelling in the sense that a separate AR model is trained on each channel. In the case of channel-independent Wavenet and GPT2 models, we train one and the same model on all channels.

\subsection{Generation and evaluation}
\label{ssec:gen_eval}

We evaluate models in both \emph{forecasting} and \emph{free-running generation} settings. Tokenized models (WaveNet and GPT2) generate by sampling tokens autoregressively (top-$p$ sampling \citep{Holtzman:2020}); AR models generate by recursively filtering noise. Since one-step prediction accuracy can be weakly informative for long-horizon behaviour, we emphasize generation-based metrics:
\begin{itemize}\itemsep0.2em
    \item \textbf{Spectral fidelity:} power spectral density (PSD) comparisons.
    \item \textbf{Multivariate dynamics:} summary statistics of a 12-state time-domain embedding HMM fit to generated vs real multichannel time series \citep{vidaurre2018spontaneous}.
    \item \textbf{Task-evoked structure:} trial-averaged evoked responses under task conditioning.
    \item \textbf{Downstream utility:} decoding performance when training/fine-tuning classifiers on simulated trials.
\end{itemize}
Implementation details for HMM fitting and decoding are provided in Appendix~\ref{ssec:appendix_eval_details}.

\section{Experimental Setup}
\label{sec:experiments}

\textbf{Dataset.} We use the 15-subject continuous MEG dataset of \citet{cichy2016comparison} (118 visual images, 30 trials/image). We apply standard preprocessing (filtering, ICA artifact rejection) and downsample to 100~Hz; details are in Appendix~\ref{ssec:appendix_data_preproc}. For within-subject trainings we create non-overlapping train/validation/test splits by holding out 4 trials/condition for validation and 4 for test (remaining 22 for training).

\textbf{Tokenization.} Each sensor is standardized (0 mean, unit variance), clipped $(-10, 10)$, rescaled to $(-1,1)$, and quantized into $Q{=}256$ bins with $\mu$-law companding. This discretization incurs negligible loss for evoked and decoding analyses (Appendix~\ref{ssec:appendix_tokenization_quality}).

\textbf{Models and training.} We compare a per-channel linear AR(255) baseline (order matched to the deep models' receptive field) with \texttt{WFC}, \texttt{WFCM}, and \texttt{GPT2MEG}. Unless otherwise stated, generative analyses are reported for a representative subject (single-subject training) to enable rapid iteration, and we additionally evaluate multi-subject scaling with \texttt{GPT2MEG-group} (15 subjects). \texttt{WFC} uses a receptive field of 255 samples via two stacks of dilated-convolution blocks; \texttt{GPT2MEG} uses 12 decoder layers, 12 heads, and $d_\mathrm{model}{=}96$ with variable context length (128--256). Deep models are trained with cross-entropy on quantized tokens and early stopping on validation loss; AR is trained with MSE on continuous signals. Full hyperparameters are reported in Appendix~\ref{ssec:appendix_hparams}.

\section{Results}

\subsection{ One-step prediction is weakly informative}
\label{ssec:flatgpt_results}

\begin{sloppypar}
    First, we assessed the models’ forecasting performance, i.e. the prediction accuracy of the label at the next time point. We used two different modified versions of Wavenet (\texttt{WavenetFullChannel} and \texttt{WavenetFullChannelMix}) alongside \texttt{GPT2MEG}. For comparison, we also evaluated the performance of a linear autoregressive (AR) model of order 255. For AR models we simply binned the predicted continuous output to compute accuracy and compare with other models.
\end{sloppypar}

\begin{figure}[!t]
    \centering
    \includegraphics[width=0.49\textwidth]{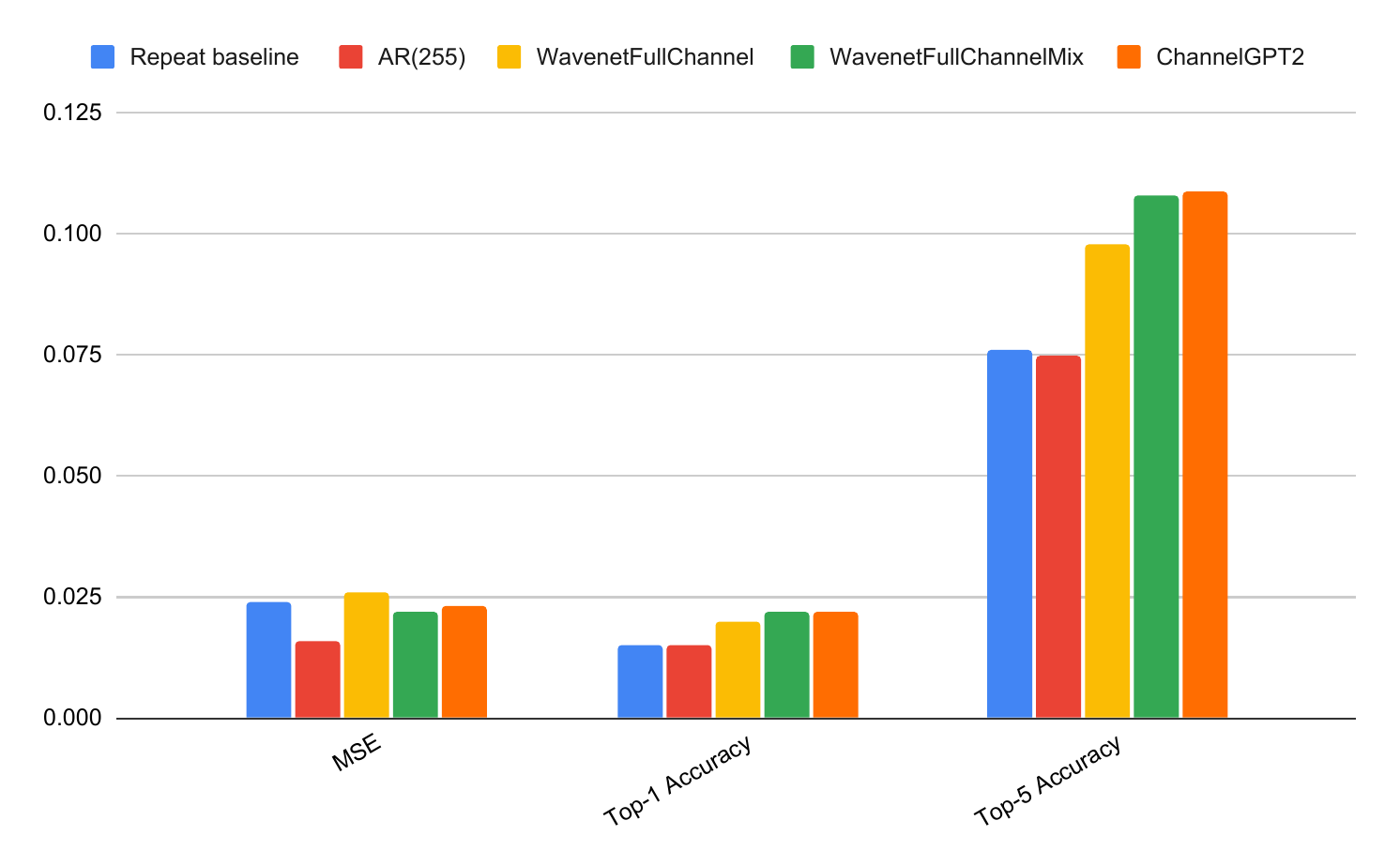}
    \caption{ \textbf{Next-timestep prediction performance across the different forecasting models.} Accuracy values are on left-out test data and are given in 0-1 units. Chance-level is $1/256$, however predicting the majority class (quantised value) is somewhat higher, since the distribution over bins is not uniform. \texttt{ChannelGPT} refers to \texttt{GPT2MEG}.}
    \label{fig:forecast_results}
\end{figure}

The results on a sample subject is shown in Figure~\ref{fig:forecast_results}. Beyond standard accuracy, we also evaluated top-5 accuracy, counting a prediction as correct if the true bin was within the 5 most probable bins. Surprisingly, all models performed only moderately better than a naive baseline of repeating the previous timestep’s value. However, as we shall investigate later, this does not necessarily reflect the richness of the structure in data recursively generated by the models.

As expected, the linear AR model had lower MSE but worse accuracy than the nonlinear models. This can be because MSE measures the distance of the prediction to the target, while accuracy is only 1 if the prediction is in the target bin. Thus, it may be that the AR model always predicts values that are slightly closer to the target, but never quite falling in the target bin.  While \texttt{WavenetFullChannel} appears to be worse, \texttt{WavenetFullChannelMix} and \texttt{GPT2MEG} have nearly identical performance. Based on these results it is inconclusive whether deep learning models improve over the linear AR model. Further, long-range structure analyses are presented in the next sections to elucidate this.

We note that the choice of sampling rate can affect forecasting performance. A higher sampling rate makes the task easier as consecutive timesteps are more correlated, however this might make the model focus on very short-range temporal dependencies and overfit to noise. We analysed forecasting performance in relation to sampling rate in Supplementary Section~\ref{ssec:sampling_rate}.

\subsection{Spectral fidelity of long-horizon generation}

A good generative model should be expected to be able to recursively generate data that looks like the real data. Here, we first assess the models’ ability to do this using the power spectra. For deep learning models we used top-p sampling with $p=80\%$ (unless otherwise noted in the figure caption) to recursively generate data. We generated 3600 seconds with all models. For models that have task-conditioning (all except AR(255)) we use the task label timeseries from the training set. The models in this section were trained on a single sample subject, containing about 1.5 hours of data downsampled to 100 Hz across 306 channels.

\begin{figure}[!t]
\centering
\begin{subfigure}{0.24\textwidth}
  \centering
  \includegraphics[width=1.0\linewidth]{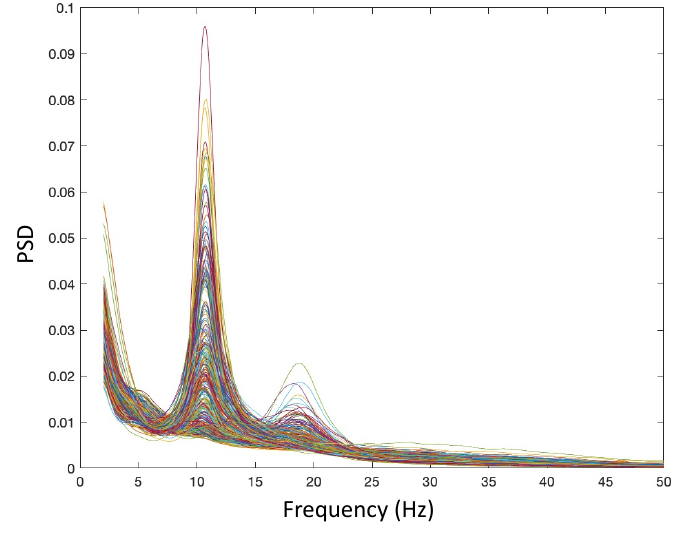}
  \caption{Data}
  \label{fig:data_psd}
\end{subfigure}%
\begin{subfigure}{0.24\textwidth}
  \centering
  \includegraphics[width=1.0\linewidth]{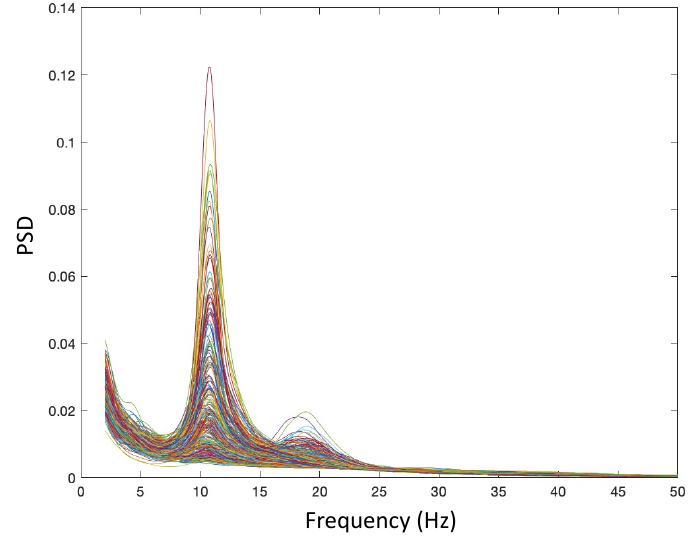}
  \caption{AR(255)}
  \label{fig:ar_psd}
\end{subfigure}
\begin{subfigure}{0.24\textwidth}
  \centering
  \includegraphics[width=1.0\linewidth]{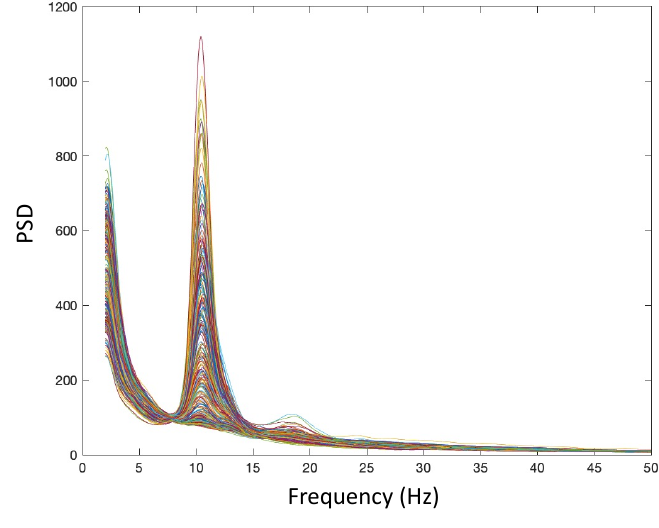}
  \caption{\texttt{WFC}}
  \label{fig:wavenefulltchannel_psd}
\end{subfigure}%
\begin{subfigure}{0.24\textwidth}
  \centering
  \includegraphics[width=1.0\linewidth]{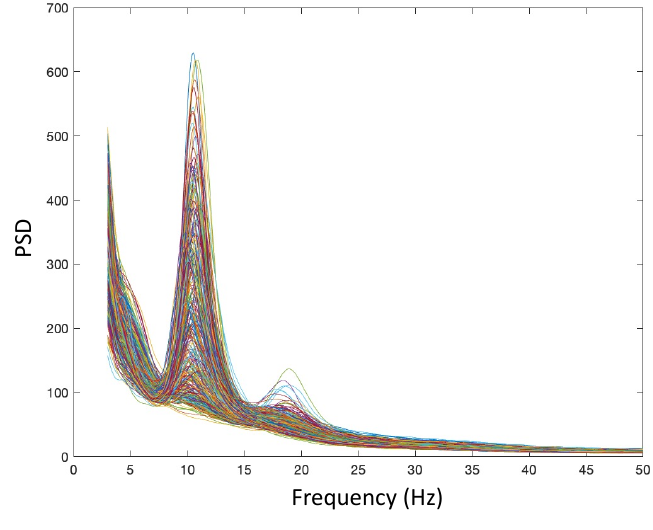}
  \caption{\texttt{GPT2MEG}}
  \label{fig:gpt2_psd}
\end{subfigure}
\caption{\textbf{PSD comparison} between real MEG (a) and long-horizon generations from different models (b--d) for a representative subject. Each line is a sensor/channel. Additional \texttt{WFCM} sampling results, generally producing less realistic spectra, are in Appendix~\ref{ssec:appendix_psd_wfcm}.}
\label{fig:psd_compare1}
\end{figure}

Generated token sequences are first de-tokenised and then the power spectral density (PSD) is computed on the continuous data. Figure~\ref{fig:psd_compare1} compares the PSD of the generated data across our models. Qualitatively, it is clear that AR(255) reproduces PSDs that match best with those computed directly on the MEG data, while \texttt{WavenetFullChannel} and \texttt{GPT2MEG} are not far behind. However, a more detailed analysis on HMM power spectra (Figure~\ref{fig:hmm_psd} in the Appendix) shows that the heterogeinity of frequency content is not capture by the AR model. Thus, while average PSD across time may look great, it cannot be used as a single evaluation point. All models capture the characteristic $1/f$ shape, and peaks at 10 and 19 Hz, likely related to alpha and beta band activity. Notably, \texttt{WavenetFullChannel} has reduced power at the 19 Hz peak, which could indicate issues in capturing higher frequency dynamics. Generated PSDs of group-level models trained on all subjects (Section \ref{ssec:group_forecasting}) were equally similar to the real data (plots not shown).

\subsection{Transformer generation better matches multivariate dynamics}

Next, we assessed how well the fitted models can recursively generate data with the same spatial, temporal and spectral multi-channel characteristics as real MEG data. Hidden Markov Models (HMMs) are an established way for doing unsupervised discovery of multi-channel dynamics in real neuroimaging data, and have been used to characterise the spatial, temporal and spectral characteristics of brain networks in MEG data \citep{rabiner1989tutorial, vidaurre2018discovering}. All results in this section are obtained using a single sample subject.

\begin{figure}[!t]
    \centering
    \includegraphics[width=0.49\textwidth]{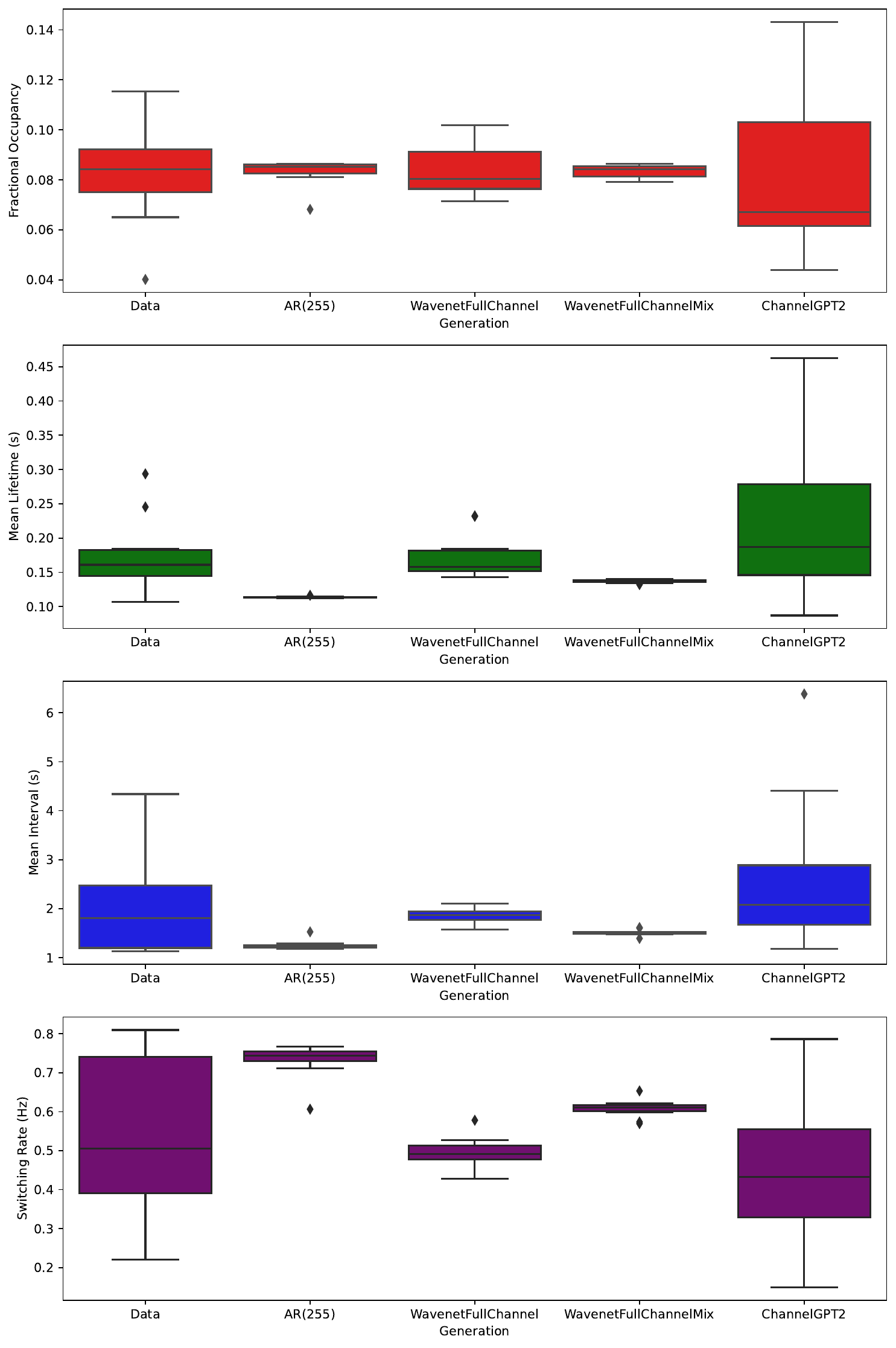}
    \caption{\textbf{Distributions of dynamics summary statistics across the 12 states from an HMM} inferred on real MEG multi-channel data from a single sample subject [left column], and from multi-channel data generated from 4 different forecasting models trained on the single sample subject. Four different summary statistics are shown describing the state dynamics (rows). \texttt{ChannelGPT} refers to \texttt{GPT2MEG}.}
    \label{fig:hmm_stats_all}
\end{figure}

Separately, we trained each model on a single sample subject (with condition embeddings), generated multi-channel data, and then inferred a 12-state HMM, with each state modelling the multi-channel data as a multi-variate Normal (MVN) distribution. The number of HMM states was chosen based on previous work \citep{vidaurre2018discovering}. Note that each recursively generated timeseries will be different, making detailed comparisons of the HMM state time courses across models meaningless. However, we can still consider broader differences in the generated dynamics, e.g. the speed of state switching. Note that since a separate HMM is trained each time, the states are not automatically matched between models or with the real data.

We extracted four summary statistics on the different inferred state timecourses and compared their distributions over states. These are shown across models in Figure~\ref{fig:hmm_stats_all}, alongside those for an HMM trained on the real multi-channel MEG data. Across the four summary statistics we can see that the real data has high variance in the distribution over states. AR(255) and \texttt{WavenetFullChannelMix} fail to produce data with variable state statistics, and even the mean over states is not captured well. \texttt{WavenetFullChannel} does a great job at capturing the mean of the state distributions, but still produces data with relatively invariant states. \texttt{GPT2MEG} is best at capturing the distributions across all four statistics, especially for the mean interval and switching rate. This shows that Transformer-based models can generate data that better matches the HMM-inferred dynamics of real MEG data.

\subsection{Conditioned generation reproduces task-evoked responses}
% HMM and evoked from data
% evoked response correlation topomap

\begin{figure}[!t]
  \centering
  \begin{subfigure}{0.49\textwidth}
    \centering
    \includegraphics[width=1.0\linewidth]{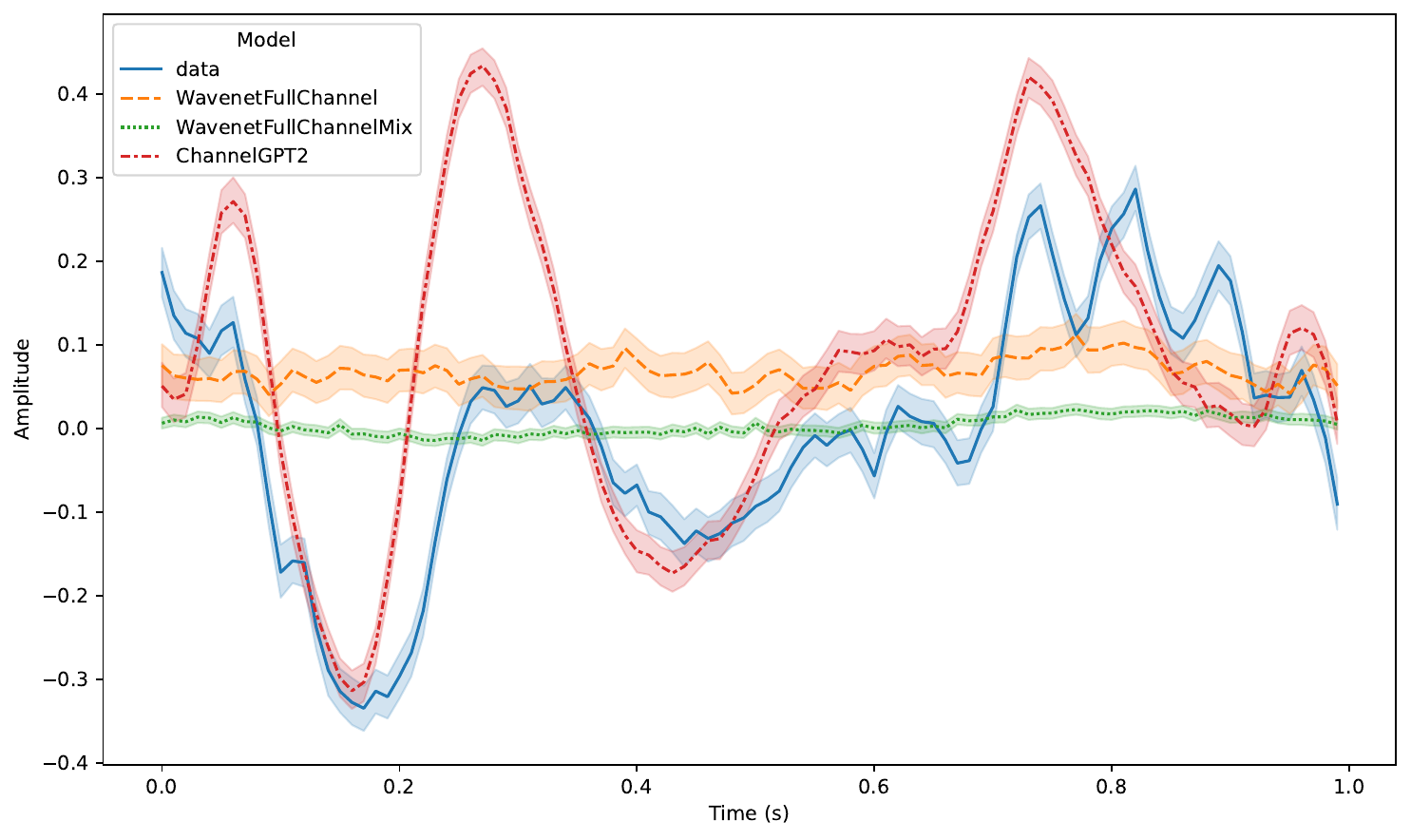}
    \caption{Frontal channel (MEG0111)}
    \label{fig:evoked_frontal_comparison}
  \end{subfigure}
  \begin{subfigure}{0.49\textwidth}
    \centering
    \includegraphics[width=1.0\linewidth]{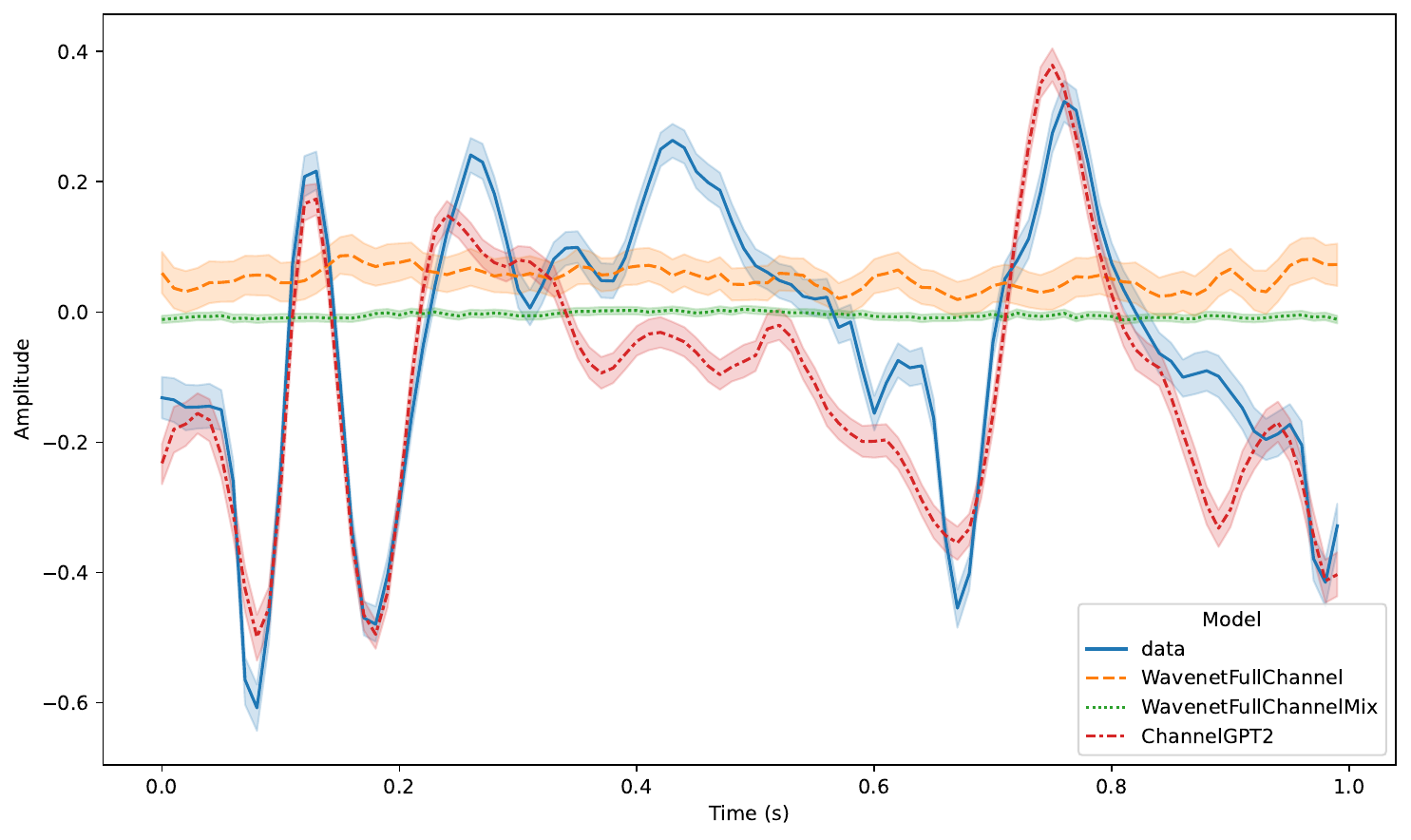}
    \caption{Visual channel (MEG2332)}
    \label{fig:evoked_visual_comparison}
  \end{subfigure}
  \caption{\textbf{Comparison of evoked timecourses of 2 channels across the task-conditioned models (via task-condition embedding)}; for real MEG data [blue line] from a single sample subject, and from data generated from the different forecasting models trained on the single sample subject. Stimulus onset is at 0 seconds and offset is at 500ms. The peak occurring after 500 ms indicates a visual response to the offset of the stimulus (removal of the image). Shading indicates the 95\% confidence interval of the trial average. \texttt{ChannelGPT} refers to \texttt{GPT2MEG}.}
  \label{fig:evoked_comparison}
  \end{figure}

The analyses in the previous section considered metrics for assessing the ability of the candidate autoregressive models to generate timeseries without requiring any a priori knowledge of the timing of brain activity. Here, we use knowledge of the experimental task timings in the \citet{cichy2016comparison} data, to provide insight into the ability of the models to generate realistic task data. 

The models in this section were trained on a single sample subject. As mentioned before, we used the task label timeseries from the training data when generating data with our models. If the models properly incorporate this conditioning, the generated data should reflect aligned task-related activity similar to real data.

By simple epoching of the generated timeseries based on the known task labels, we can compute evoked responses in the data generated by our models. We do this for all models except AR(255), as it is not able to straightforwardly include task labels in its model. To compare the shape of average evoked responses, we averaged over all epochs in both real data and the generated timeseries. This results in data of shape $\bar{\mathbf{X}} \in \mathbb{R}^{C \times T}$ where $C=306$ is the number of channels and $T=1000$ ms is the trial/epoch length.

The evoked responses across our models and the real data in a frontal and a visual channel are shown in Figure~\ref{fig:evoked_comparison}. While Wavenet models completely fail to capture the evoked time-course, \texttt{GPT2MEG} does a remarkably good job, especially in the visual channel. This is not surprising as the dataset is collected from a visual experiment, so most activity is visual. \texttt{GPT2MEG} closely matches both the amplitude and the timing of the evoked response peaks across the whole 1-second epoch. Variability across trials is also well matched (as shown by the shading in Figure~\ref{fig:evoked_comparison}). These results generalize well to multiple subjects as will be shown in Section~\ref{ssec:group_forecasting}.

Full-sensor evoked-response correlation maps and HMM-state evoked analyses are provided in Appendix~\ref{ssec:appendix_evoked}.

\subsection{Scaling to multiple subjects with subject
embeddings}
\label{ssec:group_forecasting}

Up to this point, all trainings and analyses were done on MEG data from a single sample subject. We next looked at whether combining data from multiple subjects improves modelling and generation performance. This is in line with the overall goal of training such foundational forecasting models on large datasets containing multiple subjects. Here we took a first step in exploring this by scaling \texttt{GPT2MEG}  to the 15 subjects in the \citet{cichy2016comparison} data, which we refer to as \texttt{GPT2MEG-group}. For adapting to multiple subjects and to capture variability over subjects, we used subject embeddings (see Methods). The main reason for only evaluating \texttt{GPT2MEG}  on group data is the comparatively much poorer performance of Wavenet-based models in evoked timeseries generation.

\begin{figure}[!t]
    \centering
    \includegraphics[width=0.49\textwidth]{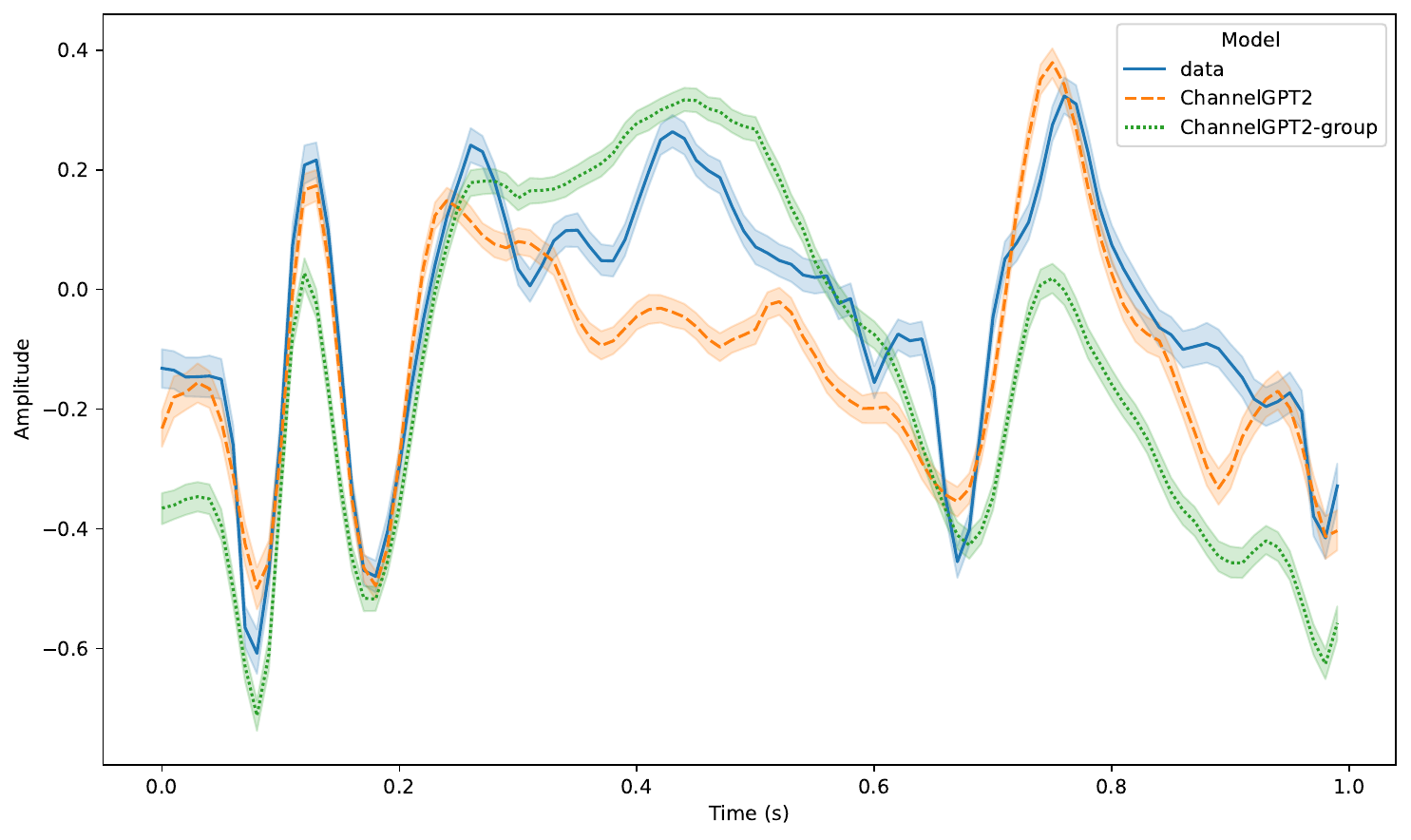}
    \caption{\textbf{Comparison of evoked responses in a visual channel (MEG2332) for a single sample subject}; using: real MEG data (blue), data generated from the \texttt{GPT2MEG} model trained on data from the single sample subject (orange), and data generated for the single sample subject (using an appropriate subject embedding) from the \texttt{GPT2MEG-group} model trained on all subjects (green). The stimulus onset is at 0 s and the stimulus offset is at 500 ms. Shading indicates 95\% confidence of the trial mean. \texttt{ChannelGPT} refers to \texttt{GPT2MEG}.}
    \label{fig:gpt_group_evokeds_visual}
\end{figure}

We were interested in whether the model generated evoked responses improved their similarity with the evoked responses from the real data, when using data from more subjects. To compare with the single-subject training we generated data using the subject embedding of that subject. The comparison of the evoked response of single-subject and group models for one 1 visual channel is shown in Figure~\ref{fig:gpt_group_evokeds_visual}. We found that generally \texttt{GPT2MEG-group} produces evoked responses that are more smoothed than the single-subject model. This is possibly because the model learns to generate data that is closer to the average statistics over subjects, and while it can adapt its generation based on the subject label, this ability is not perfect.

Evoked responses across subjects and trial-level variability in evoked HMM state activations reported in \ref{ssec:group_appendix}.

\subsection{\texttt{GPT2MEG-group} generates classifiable evoked responses}
\label{ssec:classification}

\begin{sloppypar}
We have shown that the channel-independent, Transformer-based model (\texttt{GPT2MEG}), can generate data with spatial, temporal, and spectral signatures similar to real data. We next investigated whether we can use this as a foundational model in a downstream task. Specifically, we look at the ability of \texttt{GPT2MEG} to aid in the decoding of experimental task conditions in visual task dataset from \cite{cichy2016comparison}.
\end{sloppypar}

We first investigated whether the task responses generated by the \texttt{GPT2MEG} model can be classified with performance comparable to trials of real data. This also further tests how well the model captures spatiotemporal task-related activity and information. Furthermore, if similar performance can be obtained, then \texttt{GPT2MEG} could be used to simulate an arbitrarily large number of trials to potentially improve decoding of real data through pretraining on the simulated data. This is a form of transfer learning, where the decoding model, not the forecasting model (e.g. \texttt{GPT2MEG}), is transferred.

\begin{sloppypar}
First, we generated 20 trials for all 118 conditions for one sample subject, using both \texttt{GPT2MEG} trained on the sample subject and \texttt{GPT2MEG-group} trained on all subjects (with the appropriate subject embedding of the chosen sample subject). We then trained separate linear neural network models on the real data (20 trials/condition) and these generated datasets, with an appropriate 4:1 train and validation set ratio. This achieved validation accuracies of 17.6\% (real data), 1.9\% (\texttt{GPT2MEG}), and 7.2\% (\texttt{GPT2MEG-group}). In short, while the group model generates more classifiable subject-specific task-responses, it still does not reach the classification accuracy of real data. Nonetheless, this provides further evidence that \texttt{GPT2MEG-group} successfully leverages larger datasets to produce more accurate task-related activity.
\end{sloppypar}

\subsection{Transfer learning}
\label{ssec:transfer}

A key advantage of generated data is the ability to generate huge amounts of surrogate data. As in the previous section using 20 trials per condition, we generated additional datasets with 40 and 60 trials per condition using the \texttt{GPT2MEG-group} trained model. Training a decoder on these achieved validation accuracies of 7.2\% (20 trials per condition), 21.7\% (40 trials) and 44.2\% (60 trials), exhibiting linear scaling of classification performance with the amount of simulated data.

\begin{figure}[!t]
    \centering
    \includegraphics[width=0.49\textwidth]{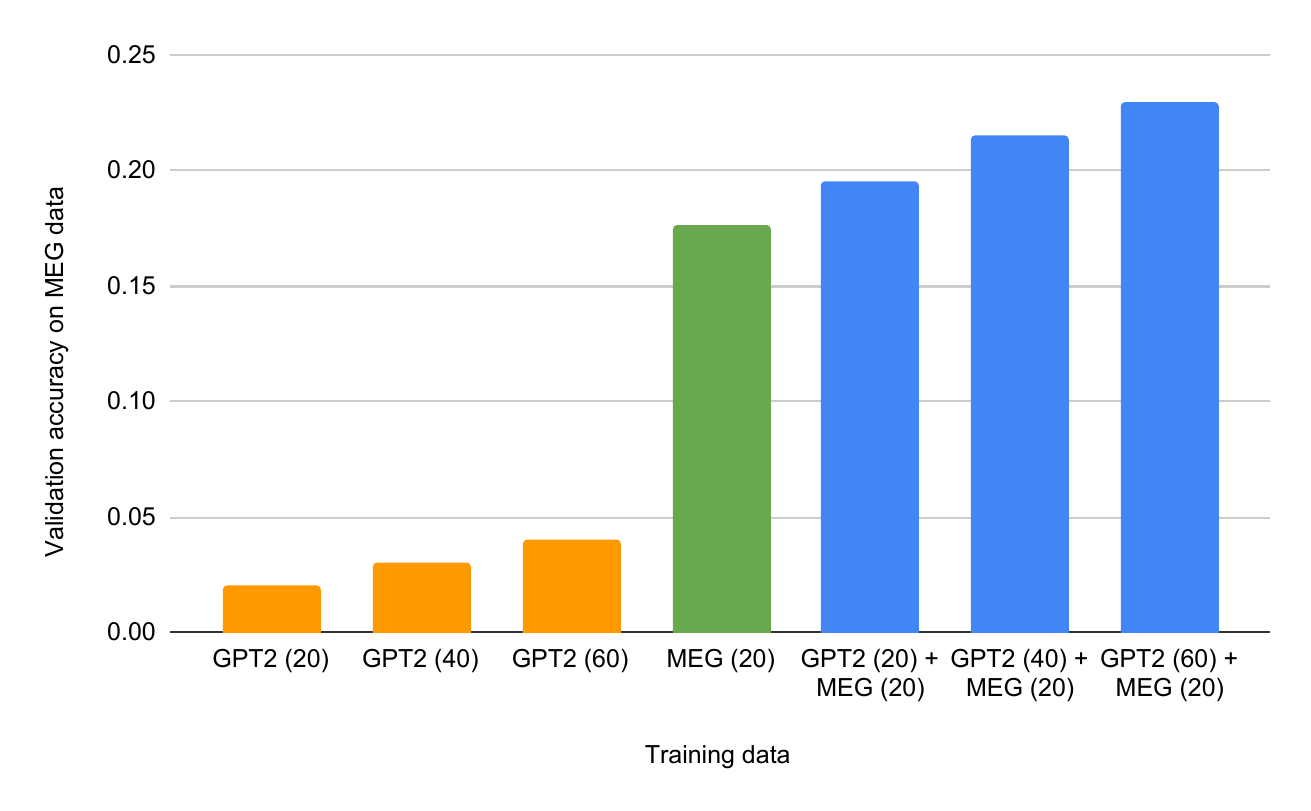}
    \caption{\textbf{Summary of the decoding accuracies of visual stimuli when using different amounts of transfer learning}. The horizontal axis represents which data the decoder was trained on. GPT2 (N) refers to the \texttt{GPT2MEG-group} generated data, while GPT2 (N) + MEG (20) is the fine-tuned decoder on the MEG data, where N is the number of trials per condition generated by \texttt{GPT2MEG-group}. The vertical axis shows the validation accuracy on the validation trials of the MEG data. Orange shows zeroshot performance, while with blue we denote the finetuned models. Chance level is $1/118$.}
    \label{fig:transfer_results}
\end{figure}

Critically, we next assessed whether this simulated data can pretrain classifiers for transfer learning. First, we took the neural network decoder pre-trained on the 20-, 40-, and 60-trial generated datasets. We then finetuned the decoder (trained it further) on the real MEG dataset (20 trials per condition), and evaluated it on separate validation trials from the MEG data. As the quantity of generated data used for pretraining increased, accuracy of the finetuned model improved rapidly. Zeroshot (no finetuning) performance on real MEG data was above chance with 2\% (20 trials per condition), 3\% (40 trials), and 4\% (60 trials) accuracy. Final accuracies after finetuning were 19.5\% (20 trials), 21.5\% (40 trials), and 23\% (60 trials). Thus, each additional 20 \texttt{GPT2MEG-group} model-generated trials per condition improved final decoding by 2\% on the real MEG trials. These results are summarised in Figure~\ref{fig:transfer_results}.

\subsection{Ablation experiments}
We performed ablation experiments with \texttt{GPT2MEG} for a single sample subject to investigate how well it can generate task-related brain activity under varied conditions without further training. In Appendix~\ref{ssec:diff_trial_length} we show that it can accurately generate evoked activity for various stimulus lengths, while being trained only on the data with the original stimulus length.

We performed two experiments to determine whether \texttt{GPT2MEG} relies solely on timing information or also utilises the semantic content of the condition labels. First, we trained a model (\texttt{GPT2MEG-randomlabel}) where the condition labels were shuffled randomly during training, breaking the semantic alignment between labels and evoked responses. Second, we trained a model (\texttt{GPT2MEG-1label}) using a single condition label for all trials. This tests whether the model cheats by learning an average evoked response instead of adapting to each task-condition.

\begin{figure}[!t]
  \centering
  \begin{subfigure}{0.16\textwidth}
    \centering
    \includegraphics[width=1.0\linewidth]{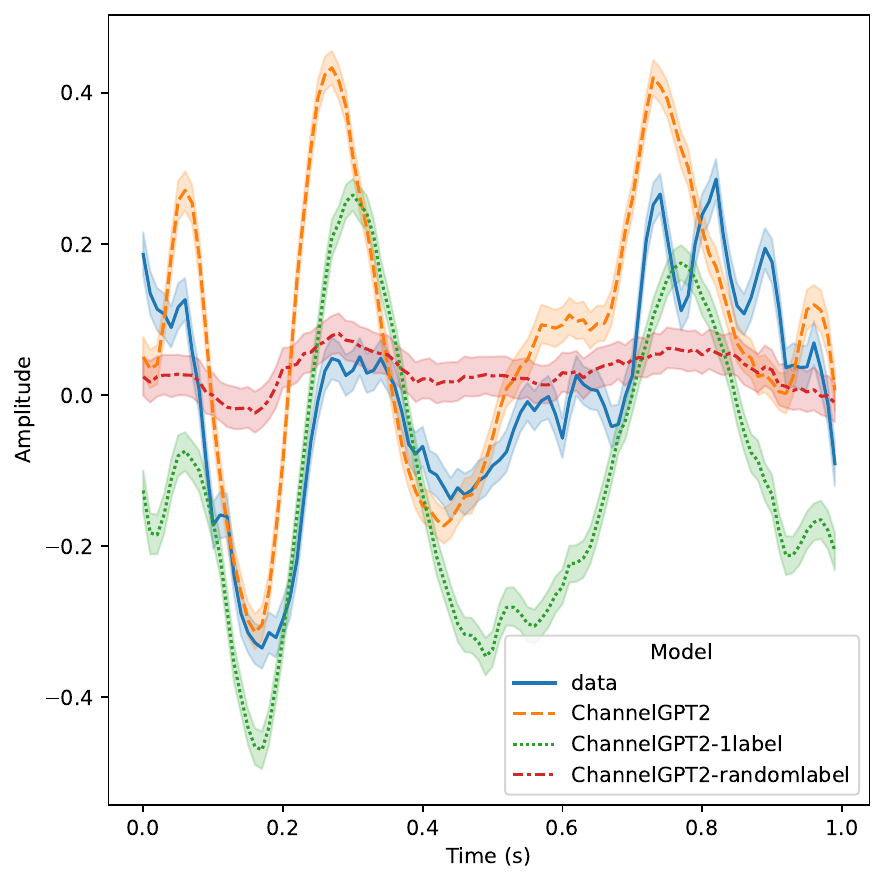}
    \caption{Ch. MEG0111}
    \label{fig:evoked_random_label_0}
  \end{subfigure}%
  \begin{subfigure}{0.16\textwidth}
    \centering
    \includegraphics[width=1.0\linewidth]{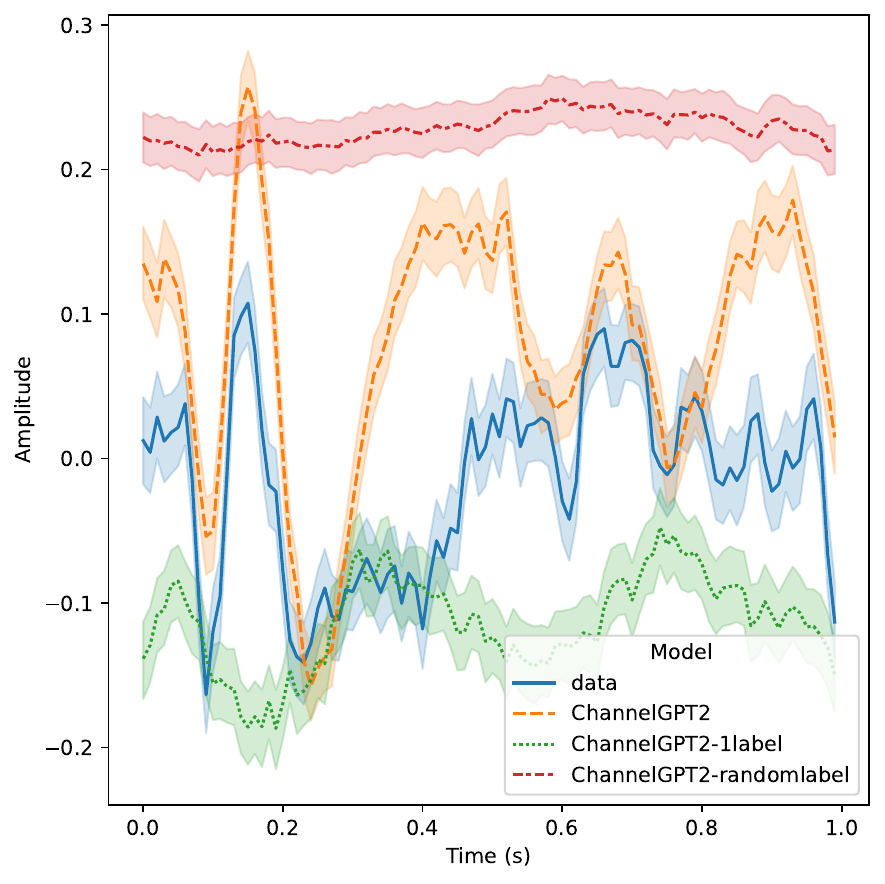}
    \caption{Ch. MEG1011}
    \label{fig:evoked_random_label_102}
  \end{subfigure}%
  \begin{subfigure}{0.16\textwidth}
    \centering
    \includegraphics[width=1.0\linewidth]{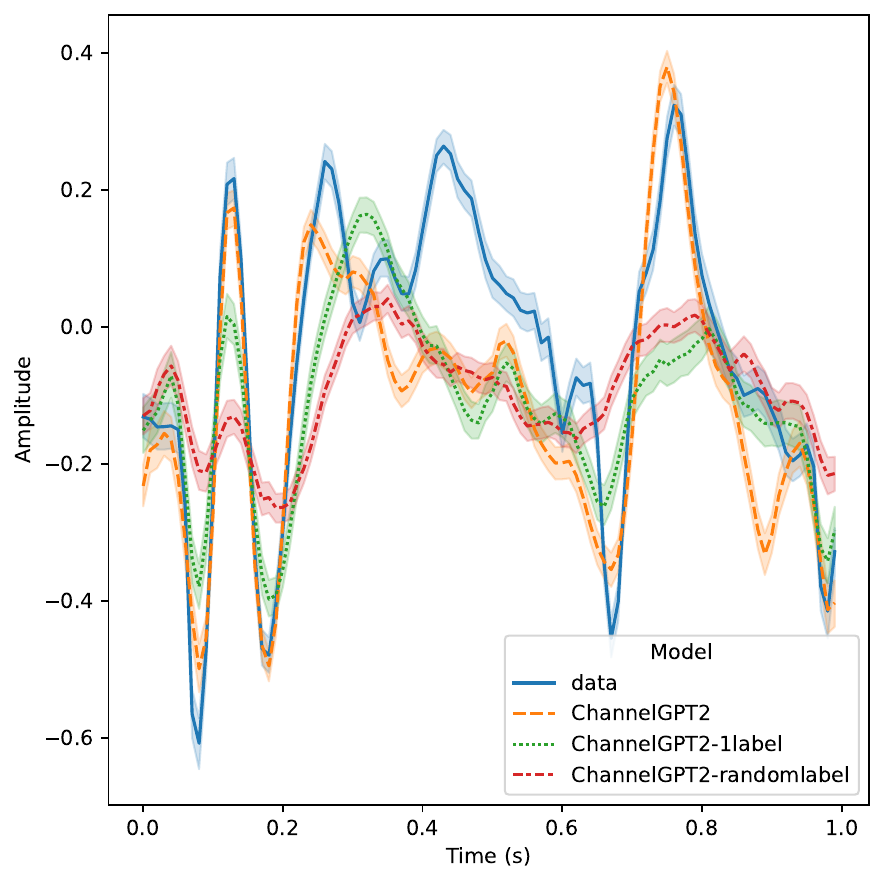}
    \caption{Ch. MEG2332}
    \label{fig:evoked_random_label_265}
  \end{subfigure}
  \caption{\textbf{Evoked responses for \texttt{GPT2MEG} models trained with shuffled or single condition labels}, indicating reliance on semantic content. Three representative channels are presented for a single sample subject. MEG0111 is anterior-left, MEG1011 is anterior-central, and MEG2332 is posterior-central. See main text for an explanation of model types. Stimulus onset is at 0 seconds, with stimulus offset at 0.5 seconds. \texttt{ChannelGPT} refers to \texttt{GPT2MEG}.}
  \label{fig:evoked_random_label}
  \end{figure}

\begin{sloppypar}
As evident in Figure \ref{fig:evoked_random_label}, models with either with shuffled or single condition labels typically failed to generate distinct evoked responses for different semantic conditions. This demonstrates that \texttt{GPT2MEG} leverages both timing and semantic information in the conditioning labels, rather than simply learning a stereotyped temporal template.  Quantitatively, evoked response correlation with real data dropped to 44\% and 56\% for \texttt{GPT2MEG-randomlabel} and \texttt{GPT2MEG-1label}, respectively, compared to 74\% for the full \texttt{GPT2MEG}.
\end{sloppypar}

We further ablated the channel and condition embeddings and analyzed the learned channel-embedding geometry; these results indicate both embeddings are important and that spatial structure emerges in the channel embeddings (Appendix~\ref{ssec:appendix_ablation_embeddings}).

\section{Discussion}

We introduced \texttt{GPT2MEG}, an autoregressive model for MEG trained by next-step prediction on unlabelled recordings. A simple discretization of MEG into tokens, together with channel, condition, and subject embeddings, allows a GPT-2--style Transformer (\texttt{GPT2MEG}) to generate realistic long-horizon dynamics and task-evoked responses. Across spectral, HMM-based dynamical, and evoked-response evaluations, Transformers outperform WaveNet variants and a linear AR baseline, despite similar one-step forecasting accuracy.

\subsection{Limitations and future work}

A core limitation of the channel-independent GPT2 model is that there is no direct leveraging of cross-channel information for each sensor prediction. We think that maintaining the innate inductive biases of Transformers, which emphasise 1D sequence modelling on embeddings of discrete tokens, is paramount. Different architectures, or more data, may enable proper utilisation of cross-channel dependencies. We tried various other approaches to mixing channel information beyond those reported, without success. For example, in the Wavenet model, we incorporated all channels in the input by concatenating embeddings, and for the GPT2 models, we tried mixing channels with convolutions. We tried concatenating the output of each channel and then predicting from this shared output using a different projection for each channel. We also attempted to increase receptive field, dropout, and model size. 

Some of our findings substantiated that predicting the next timestep's distribution (through the cross-entropy loss) may not serve as a robust measure of modelling performance. Future research should contemplate adopting multi-timestep or contrastive loss frameworks. A plausible strategy could involve deploying the VQ-VAE model across both channel and temporal dimensions, aiming to distill a coarser sequence of discrete tokens. Nevertheless, any quantisation-centric approach must carefully consider reconstruction error. We posit that a significant portion of the signal dynamics should be entrusted to the Transformer, given its adeptness in capturing complex dynamics.

A constraint in our modelling approach is its reliance on categorical task stimuli labels. Such an approach, while effective in our context, does not readily lend itself to scalability across diverse tasks and datasets. Robust representations tailored for various stimulus modalities, ranging from images to audio, can serve as conditioning embeddings. As shown by \citet{defossez2022decoding}, tools such as wav2vec \citep{baevski2020wav2vec} can be leveraged for this.

Future work should explore more flexible conditioning, study different self-supervised and transfer learning frameworks, and critically, apply similar analyses when scaling up across diverse, large electrophysiology datasets.

% Acknowledgements should only appear in the accepted version.
\section*{Acknowledgements}
This research was supported by the NIHR Oxford Health Biomedical Research Centre. The views expressed are those of the author(s) and not necessarily those of the NIHR or the Department of Health and Social Care. RC is supported by a Wellcome Centre Integrative Neuroimaging Studentship. MVE's research is supported by the Wellcome Trust (215573/Z/19/Z). OPJ is supported by the UK MRC (MR/X00757X/1). MWW's research is supported by the Wellcome Trust (106183/Z/14/Z, 215573/Z/19/Z), the New Therapeutics in Alzheimer’s Diseases (NTAD) study supported by UK MRC and the Dementia Platform UK (RG94383/RG89702) and the EU-project euSNN (MSCA-ITN H2020-860563). The Wellcome Centre for Integrative Neuroimaging is supported by core funding from the Wellcome Trust (203139/Z/16/Z).

\section*{Impact Statement}

This work develops autoregressive generative models for MEG that can simulate realistic task-evoked neural time series. Potential positive impacts include data augmentation for data-scarce decoding settings and reproducible benchmarking of analysis pipelines. Potential risks include misuse of synthetic data as a substitute for real measurements, amplification of biases present in the training data (e.g.\ demographic, acquisition, or preprocessing biases), and over-interpretation of generated signals as mechanistic evidence. We mitigate these risks by evaluating generation with multiple neuroscience-informed metrics. Future work should study calibration, dataset diversity, and safeguards for clinical and BCI applications.

\bibliography{ml}
\bibliographystyle{icml2026}

%%%%%%%%%%%%%%%%%%%%%%%%%%%%%%%%%%%%%%%%%%%%%%%%%%%%%%%%%%%%%%%%%%%%%%%%%%%%%%%
%%%%%%%%%%%%%%%%%%%%%%%%%%%%%%%%%%%%%%%%%%%%%%%%%%%%%%%%%%%%%%%%%%%%%%%%%%%%%%%
% APPENDIX
%%%%%%%%%%%%%%%%%%%%%%%%%%%%%%%%%%%%%%%%%%%%%%%%%%%%%%%%%%%%%%%%%%%%%%%%%%%%%%%
%%%%%%%%%%%%%%%%%%%%%%%%%%%%%%%%%%%%%%%%%%%%%%%%%%%%%%%%%%%%%%%%%%%%%%%%%%%%%%%
\newpage
\appendix
\onecolumn

\section{Appendix}

\subsection{Multi-channel Wavenet}
\label{ssec:appendix_wavenet_old}

Here we describe how we adapted the Wavenet architecture \citep{oord2016wavenet} for electrophysiological data. Wavenet models the conditional probability of each time sample given all preceding samples autoregressively:

\begin{equation}
p(\mathbf{X}) = \prod_{t=1}^{T} p(\mathbf{x}_t | \mathbf{x}_1, ..., \mathbf{x}_{t-1})
\end{equation}

where $\mathbf{x}_t$ is the sample at time $t$ and $T$ is the total sequence length. Throughout this paper we use tokenisation and quantisation interchangeably. Both have the aim of discretising a continuous quantity into a finite set of distinct bins/levels/tokens.

In the original paper, the audio waveform is tokenised using a quantisation to 8 bits following a $\mu$-law companding transform \citep{lewis1997law}:

\begin{equation}
    f(\mathbf{x}_t) = \mathrm{sign}(\mathbf{x}_t)\frac{\ln(1 + \mu |\mathbf{x}_t|)}{\ln(1 + \mu)}
\end{equation}

where $\mu$ controls the number of quantisation levels, set to 255 as in the original Wavenet. $f(.)$ is applied to each value of $\mathbf{x}_t$ independently. This nonlinear transformation improves reconstruction versus uniform quantisation of the raw input, as it skews the distribution such that more levels are allocated to smaller magnitudes. For MEG data, we observe similar benefits when applying this transform prior to quantisation. Note that the input must be scaled to $(-1,1)$ first, and clipping outliers above some threshold helps ensure a more uniform mapping.

\begin{sloppypar}
When adapting Wavenet to M/EEG, a key challenge is the multi-channel nature of the data. We devise two versions: \texttt{WavenetFullChannel} as univariate, and \texttt{WavenetFullChannelMix} as multivariate. In both, each channel is transformed and tokenised independently to form the input to the models.
\end{sloppypar}

In \texttt{WavenetFullChannel}, we first apply an embedding layer to the tokenised data, learned separately per channel. To be clear in this univariate approach the same model is applied to each channel. However, a different embedding layer is learned for each channel, meaning that for example the quantised value of 0.42 in channel x will have a different vector representation than in channel y. This helps the model differentiate between channels.

The embedding operation is given below:

\begin{align}
\forall c \in {1, 2, \dots, C}: \mathbf{X}_{e}^{(c)} &= \mathbf{W}^{(c)} \mathbf{X}^{(c)}  \\
    \mathbf{H}_0 &= \mathrm{Concatenate}(\mathbf{X}_{e}^{(1)}, \mathbf{X}_{e}^{(2)}, \dots, \mathbf{X}_{e}^{(C)})
\end{align}

Here, $\mathbf{X}^{(c)} \in \mathbb{R}^{Q \times T}$ is the tokenised one-hot input and $\mathbf{W}^{(c)} \in \mathbb{R}^{E \times Q}$ is the embedding layer of channel $c$ mapping tokens $Q$ to embeddings of size $E$. $\mathrm{Concatenate}$ concatenates along the channel dimension.

$\mathbf{H}_0 \in \mathbb{R}^{C \times E \times T}$ is the resulting input to Wavenet with $C$ as the batch dimension. Thus, the same model is applied independently to each channel in parallel. At output, a distribution is predicted simultaneously for each channel at $T+1$. The model is optimised to accurately predict all channels.

\texttt{WavenetFullChannelMix} includes an extra linear layer after summing the skip representations to mix information across the channel dimension:

\begin{align}
    \mathbf{S} &= \sum_{l=1}^L\mathbf{S}^{(l)} \\
    \mathbf{S} &= \mathbf{S}\mathrm{.permute}(1, 2, 0) \\
    \mathbf{S}_{out} &= \mathbf{S}\mathbf{W}_m
\end{align}

where $\mathbf{W}_m \in \mathbb{R}^{C \times C}$ is the mixing weight matrix, and $\mathbf{S}^{(l)}$ is the output of the skip connection at layer $l$. The permutation is needed to apply the projection to the appropriate channel dimension. After this $\mathbf{S}_{out}$ is permuted back to the original dimension order and the rest proceeds identically to \texttt{WavenetFullChannel}.

In the original Wavenet, audio generation can be conditioned on additional inputs through embedding-based global conditioning or time-aligned local conditioning. For some experiments, we augment the model with local features of task stimuli or subject labels, first embedded into continuous vectors:

\begin{align}
    \mathbf{H}_y &= \mathbf{Y}\mathbf{W}_y \\
    \mathbf{H}_o &= \mathbf{O}\mathbf{W}_o \\
    \mathbf{H}_{c} &= \mathrm{Concatenate}(\mathbf{H}_y, \mathbf{H}_o)
\end{align}

where $\mathbf{Y} \in \mathbb{R}^{T \times N}$ contains the condition index $n \in (1, \dots, N)$ at each time point, and $\mathbf{O} \in \mathbb{R}^{T \times S}$ contains the subject index $s \in (1, \dots, S)$ at each time point $t \in (1, \dots, T)$. $\mathbf{W}_y \in \mathbb{R}^{N \times E_n}$ and $\mathbf{W}_o \in \mathbb{R}^{S \times E_s}$ are embedding matrices mapping the labels to learned continuous vectors of size $E_n$ and $E_s$, respectively. The subject index is the same across time points of the recording from the same subject. The condition index is set to the (visual) stimuli presented (e.g., one of the 118 images in \citet{cichy2016comparison}), for exactly those time points when the stimulus is on. At any other time, the task condition embedding $\mathbf{H}_y$ is set to 0.

$\mathbf{H}_{c}$ is the conditioning vector fed into Wavenet at each layer:

\begin{align}
\mathbf{Z}^{(l)} &= \tanh\left(\mathbf{W}^{(l)}_{f}*\mathbf{H}^{(l)} + \mathbf{W}^{(l)}_{c}*\mathbf{H}_{c}\right) \odot \sigma\left(\mathbf{W}^{(l)}_{g}*\mathbf{H}^{(l)} + \mathbf{W}^{(l)}_{c}*\mathbf{H}_{c}\right) 
\end{align}

where $\mathbf{W}^{(l)}_{c}$ (1x1 convolution) projects $\mathbf{H}_{c}$ before adding it to the input representation ($\mathbf{H}^{(l)}$). $\mathbf{W}^{(l)}_{f}$ is the filter convolution weight, $\mathbf{W}^{(l)}_{g}$ is the gate convolution weight, and $\mathbf{Z}^{(l)}$ is the output representation at layer $l$. $\odot$ is element-wise multiplication. This formulation conditions the prediction on both past brain activity and stimuli:

\begin{align}
    p(\mathbf{X}|\mathbf{Y}, \mathbf{O}) = \prod_{t=1}^{T} p(\mathbf{x}_t | \mathbf{x}_1, ..., \mathbf{x}_{t-1}, \mathbf{y}_{1}, ..., \mathbf{y}_{t-1}, \mathbf{o}_{1}, ..., \mathbf{o}_{t-1})
\end{align}

In single-subject models we only use the task labels $\mathbf{Y}$.

\subsection{Evaluation details}
\label{ssec:appendix_eval_details}
We fit 12-state time-domain embedding HMMs using osl-dynamics \citep{gohil2023osl} with 15 embeddings and PCA to 80 dimensions (sequence length 2000). Decoding uses the 4-layer linear network of \citet{csaky2023interpretable}.

\subsection{Preprocessing and data splits}
\label{ssec:appendix_data_preproc}
We bandpass filter 1--50~Hz, apply a notch filter for line noise, perform ICA artifact rejection (64 components with manual component rejection), and downsample to 100~Hz. We split the continuous recording into non-overlapping blocks corresponding to trials, holding out 4 trials/condition for validation and 4 for test.

\subsection{Tokenization quality}
\label{ssec:appendix_tokenization_quality}
We verified that de-quantized signals preserve evoked responses and decoding performance compared to raw continuous data; reconstruction error is low (less than 5\%) for both $\mu$-law and linear quantization.

\subsection{Hyperparameters}
\label{ssec:appendix_hparams}

\begin{table}[t!]
\centering
        \begin{tabular}{l|ccc}
           \bf Model  &\bf Univariate & \bf Tokenised & \bf Linear \\ \midrule
            AR(255) & yes & no & yes  \\
            \texttt{WFC} & yes & yes & no \\
            \texttt{WFCM} & no & yes & no \\
            \texttt{GPT2MEG} & yes & yes & no \\
        \end{tabular}
    \caption{\label{table:models} {\bf Summary of evaluated models.} ``Univariate'' treats channels as a batch dimension during training; ``Linear'' denotes linear dynamics in the model class.}
\end{table}

We match receptive fields across deep models (255 samples). WaveNet variants use two dilation stacks (7 layers each) with early stopping. We set dilation and residual channels to 128, and skip channels to 512. \texttt{GPT2MEG} uses 12 layers and 12 attention heads with embedding size 96 for single-subject models and 240 for group models; optimization uses Adam \citep{Kingma:2014} with early stopping. Batch size is set to the number of channels, so 1 full example for channel-independent models. Please see Table~\ref{table:models} for a summary on model variants.

\label{ssec:appendix_psd_wfcm}
\begin{figure}[!t]
  \centering
\begin{subfigure}{0.49\textwidth}
  \centering
  \includegraphics[width=1.0\linewidth]{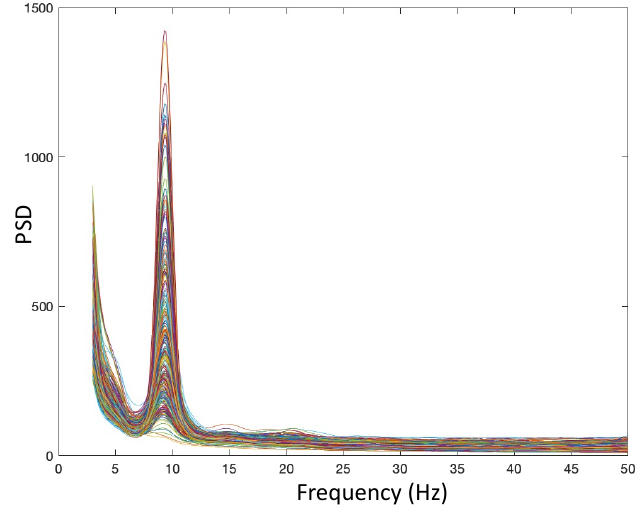}
  \caption{\texttt{WavenetFullChannelMix} p=0.72}
  \label{fig:wavenetfullchannelmix_p72_psd}
\end{subfigure}%
\begin{subfigure}{0.49\textwidth}
  \centering
  \includegraphics[width=1.0\linewidth]{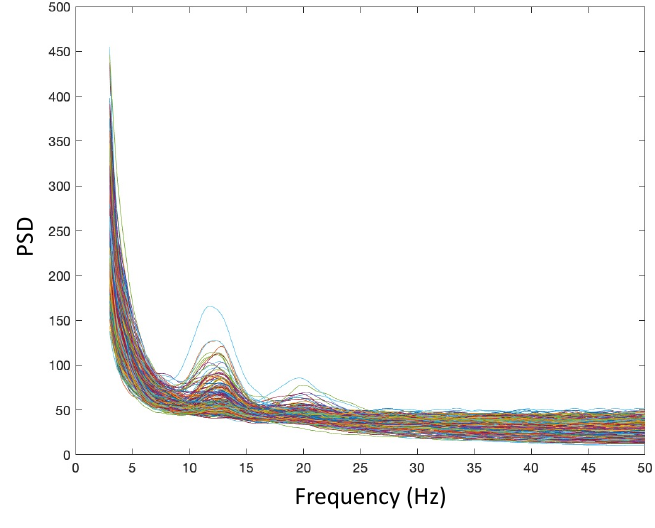}
  \caption{\texttt{WavenetFullChannelMix} p=0.80}
  \label{fig:wavenetfullchannelmix_p80_psd}
\end{subfigure}
  \caption{\textbf{PSDs for \texttt{WavenetFullChannelMix} at different top-$p$ values}; (a) $p=0.72$ and (b) $p=0.80$ for a model trained on a single subject. Each line represents a different MEG sensor/channel.}
  \label{fig:psd_wfcm}
\end{figure}

\subsection{State-specific PSDs of HMM states}
\label{ssec:appendix_hmm_psd}
\begin{figure}[!t]
  \centering
  \begin{subfigure}{0.2\textwidth}
    \centering
    \includegraphics[width=1.0\linewidth]{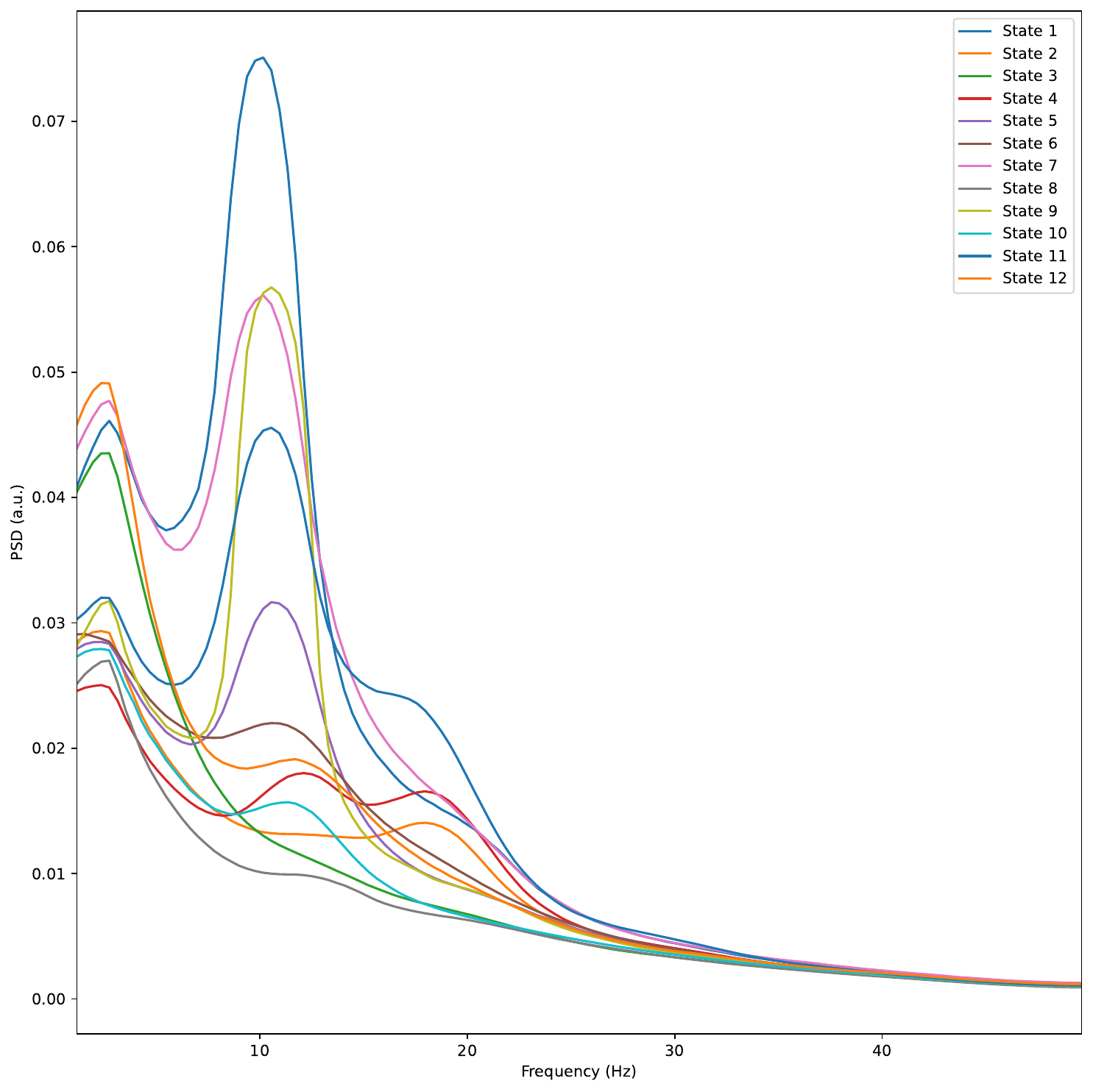}
    \caption{Data}
    \label{fig:data_hmm_psd}
  \end{subfigure}%
  \begin{subfigure}{0.2\textwidth}
    \centering
    \includegraphics[width=1.0\linewidth]{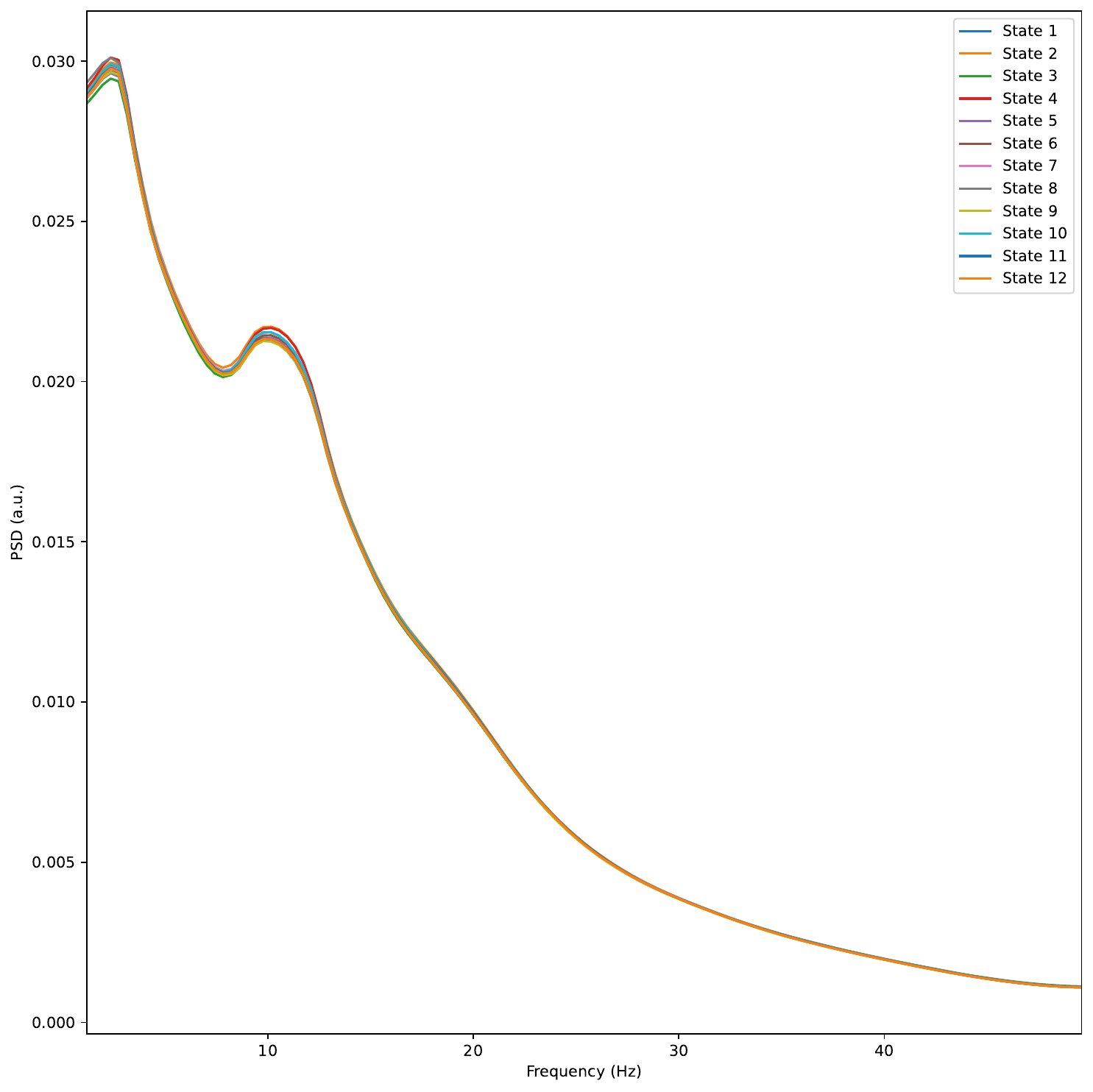}
    \caption{AR(255)}
    \label{fig:ar_hmm_psd}
  \end{subfigure}%
  \begin{subfigure}{0.2\textwidth}
    \centering
    \includegraphics[width=1.0\linewidth]{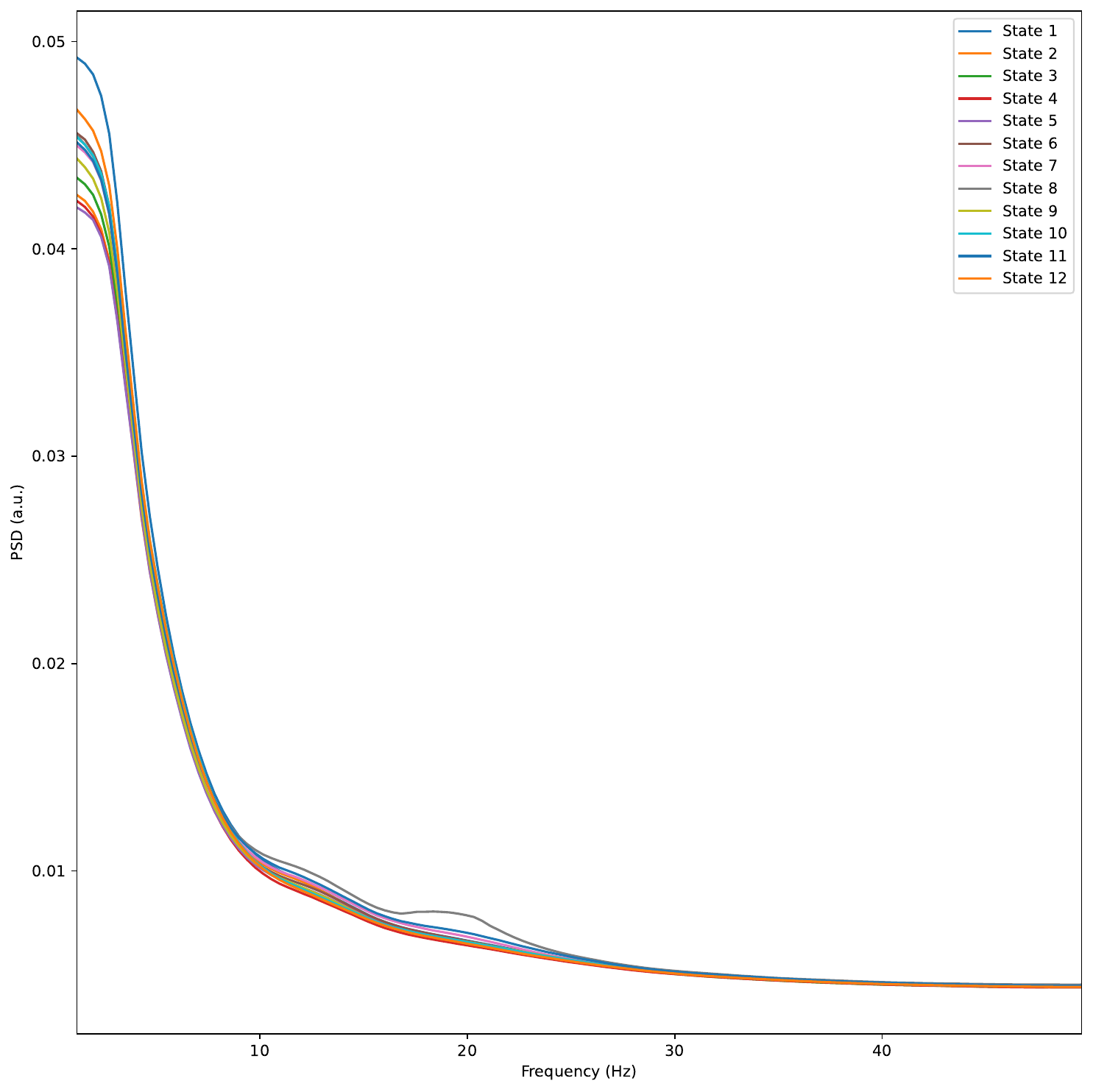}
    \caption{\texttt{WFCM}}
    \label{fig:wavenet_hmm_psd}
  \end{subfigure}%
  \begin{subfigure}{0.2\textwidth}
    \centering
    \includegraphics[width=1.0\linewidth]{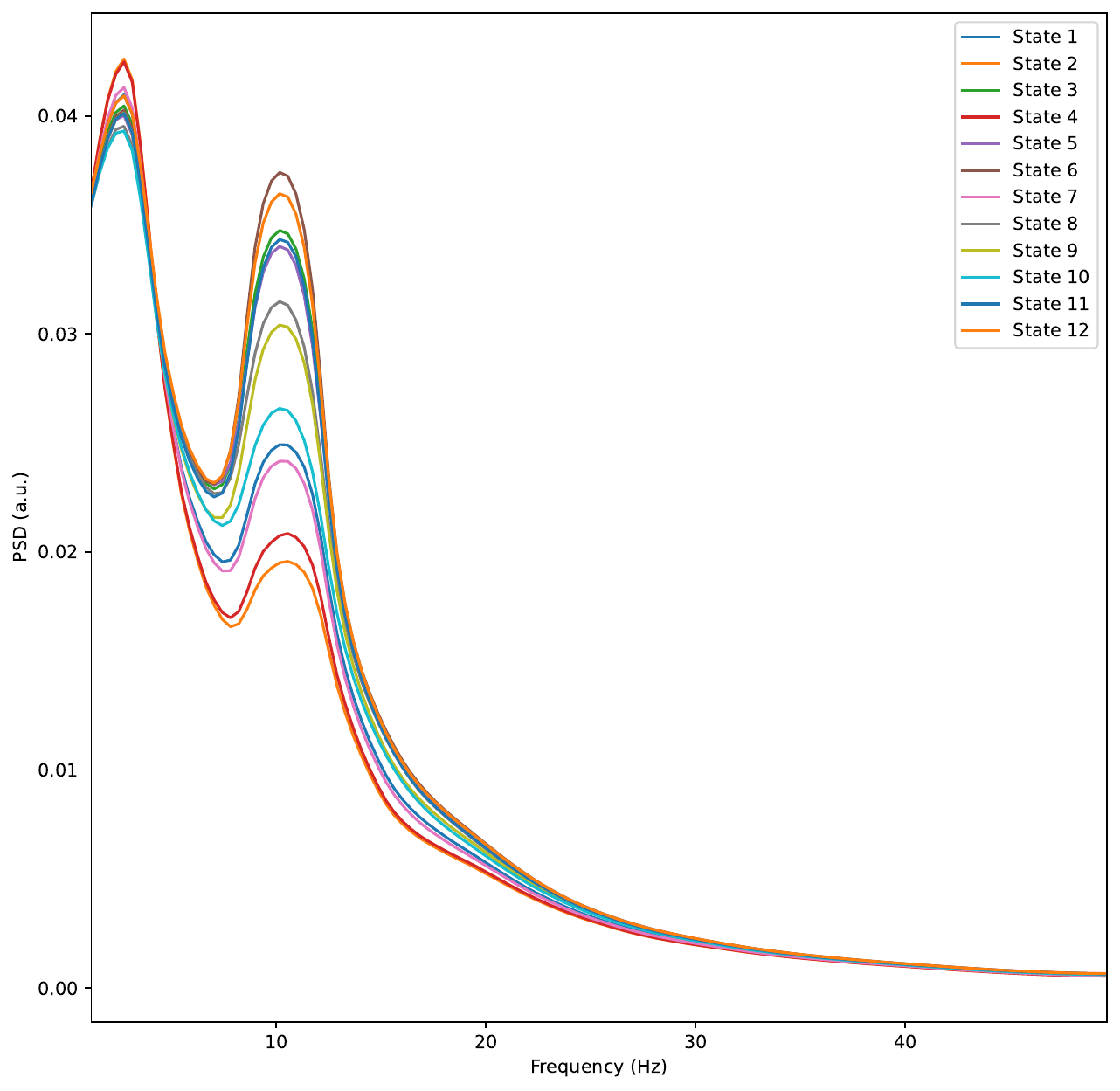}
    \caption{\texttt{WFC}}
    \label{fig:wavenetfullchannel_hmm_psd}
  \end{subfigure}%
   \begin{subfigure}{0.2\textwidth}
    \centering
    \includegraphics[width=1.0\linewidth]{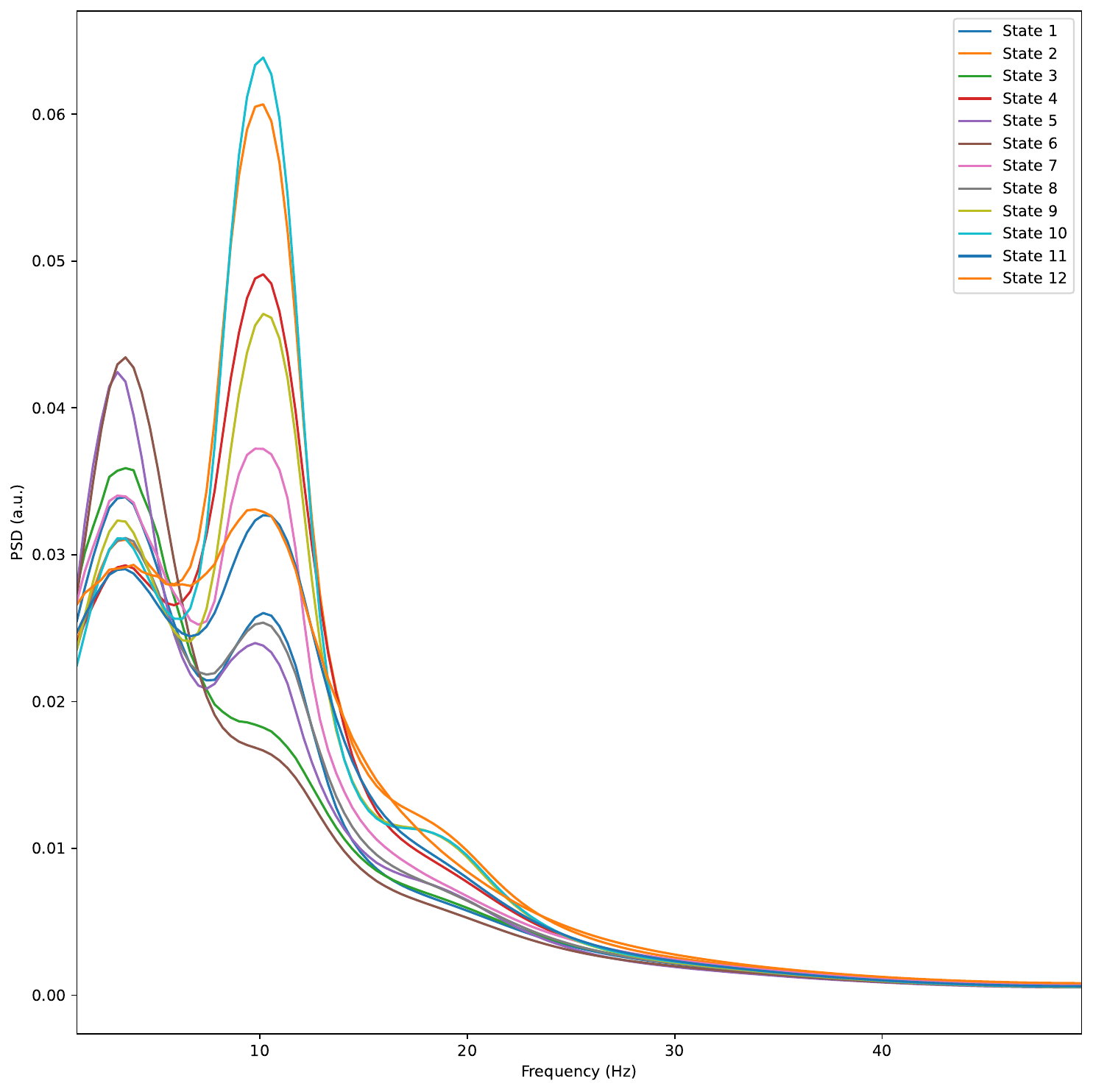}
    \caption{\texttt{GPT2MEG}}
    \label{fig:gpt_hmm_psd}
  \end{subfigure}
  \caption{\textbf{Power spectral density of HMM states}, for an HMM inferred on (a) real MEG data from a single sample subject, and (b)-(e) data generated from the different forecasting models trained on the single sample subject. \texttt{WFCM} refers to \texttt{WavenetFullChannelMix}. Each line is the PSD of a different state. Note that states are not matched across models. Horizontal axis represents frequency in Hz.}
  \label{fig:hmm_psd}
  \end{figure}

In addition to state statistics, we also computed the power spectra of each state across the timeseries. In real MEG data, different HMM states can capture oscillatory activity with specific frequencies \citep{vidaurre2018spontaneous}. The extracted power spectra from the different inferred state time courses are shown in Figure~\ref{fig:hmm_psd}. We can see that the HMM trained on the real MEG data contains many states that capture the 10 Hz peak, with fewer states having a 20 Hz peak. It is also clear that the states of the HMM fitted to the \texttt{WavenetFullChannelMix} generated timeseries do not contain these spectral peaks. While the AR(255) does contain states with a 10 Hz peak, the shape does not match the data well, and also states do not show the same variability as in real data.

In contrast, \texttt{GPT2MEG}, matches the state PSDs of the real data very well, further demonstrating the superiority of Transformer models in capturing complex neural dynamics. While \texttt{WavenetFullChannel} also improves substantially over the AR(255) power spectra, it falls short in capturing the 20 Hz peak and the heterogeneity between states observed in the real data and the generated data of \texttt{GPT2MEG}. While we have not conducted this specific analysis on multiple subjects, we see no reason why the main findings would not generalize.

\subsection{Additional evoked-response analyses}
\label{ssec:appendix_evoked}
\begin{figure}[!t]
  \centering
  \begin{subfigure}{0.16\textwidth}
    \centering
    \includegraphics[width=1.0\linewidth]{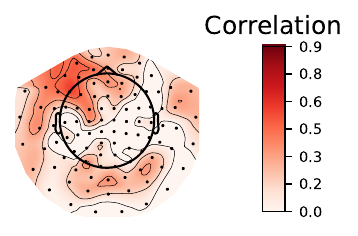}
    \caption{\texttt{WFC} mean}
    \label{fig:WavenetFullChannel_meancorr}
  \end{subfigure}%
  \begin{subfigure}{0.16\textwidth}
    \centering
    \includegraphics[width=1.0\linewidth]{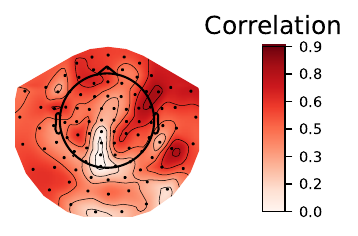}
    \caption{\texttt{WFCM} mean}
    \label{fig:WavenetFullChannelMix_meancorr}
  \end{subfigure}%
  \begin{subfigure}{0.16\textwidth}
    \centering
    \includegraphics[width=1.0\linewidth]{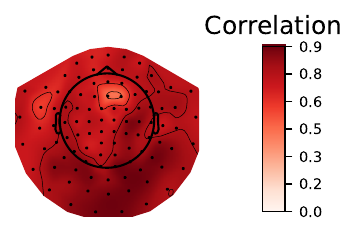}
    \caption{\texttt{GPT2MEG} mean}
    \label{fig:GPT2MEG_meancorr}
  \end{subfigure}
  \begin{subfigure}{0.16\textwidth}
    \centering
    \includegraphics[width=1.0\linewidth]{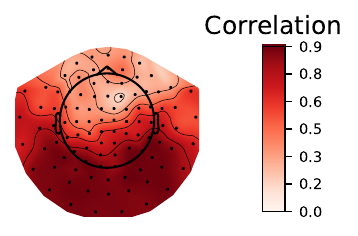}
    \caption{\texttt{WFC} var.}
    \label{fig:WavenetFullChannel_varcorr}
  \end{subfigure}%
  \begin{subfigure}{0.16\textwidth}
    \centering
    \includegraphics[width=1.0\linewidth]{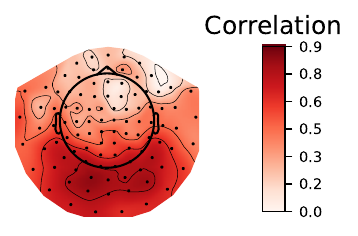}
    \caption{\texttt{WFCM} var.}
    \label{fig:WavenetFullChannelMix_varcorr}
  \end{subfigure}%
  \begin{subfigure}{0.16\textwidth}
    \centering
    \includegraphics[width=1.0\linewidth]{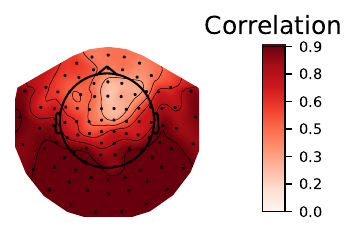}
    \caption{\texttt{GPT2MEG} var.}
    \label{fig:GPT2MEG_varcorr}
  \end{subfigure}%
  \caption{(a)-(c) \textbf{Correlation between the time-courses of the mean (over individual epochs) evoked responses}; from the real MEG data for a single sample subject and the mean evoked responses from data generated by the different forecasting models trained on the single sample subject. (d)-(e) \textbf{Correlation between the time-courses of the variance (over individual epochs) of the mean evoked responses}; from the real MEG data for a single sample subject and the mean evoked responses from data generated by the different forecasting models trained on the single sample subject. For all figures, the correlation values are visualised across sensors. \texttt{WFC} refers to \texttt{WavenetFullChannel} and \texttt{WFCM} refers to \texttt{WavenetFullChannelMix}. Darker reds indicate higher correlation.}
  \label{fig:mean_evoked_corrs}
  \end{figure}

  \begin{figure}[!t]
    \centering
    \includegraphics[width=0.8\textwidth]{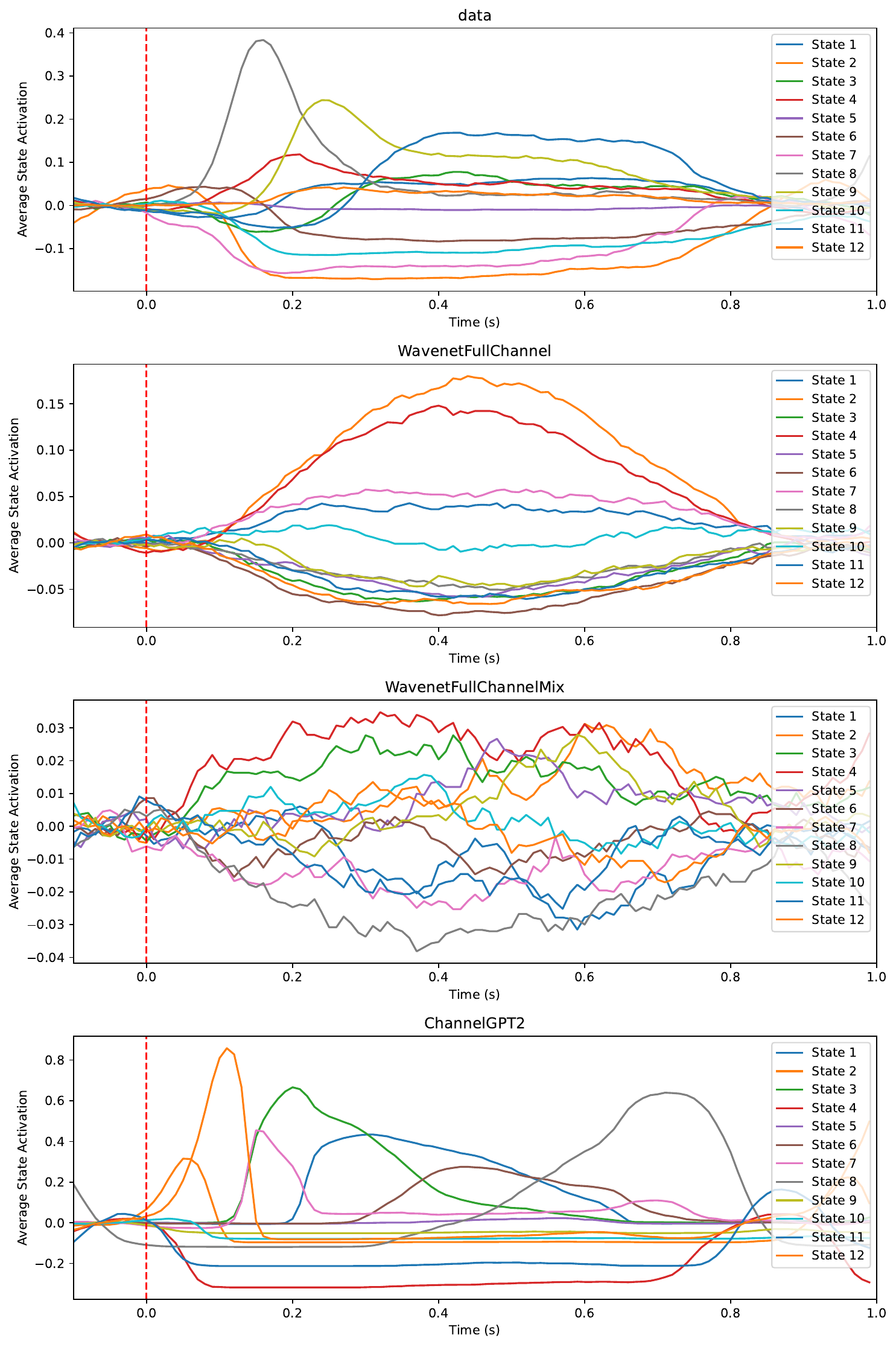}
    \caption{\textbf{Evoked response state timecourses of HMMs inferred on a single sample subject}; using real MEG data (top), and on generated data from each of our task-conditioned models trained on the single sample subject. Note that the HMM states are not matched between models. Image presentation starts at 0 seconds and ends at 0.5 seconds. \texttt{ChannelGPT} refers to \texttt{GPT2MEG}.}
    \label{fig:hmm_evokeds}
\end{figure}

To quantify the similarity between real and model generated evoked activity, we computed the correlation of the mean (across individual epochs) time-courses of the evoked response for each channel separately. Note that we averaged over the different MEG sensors (the magnetometers and gradiometers) found at the same location. The result of this is plotted in Figure~\ref{fig:mean_evoked_corrs}, allowing insights into the spatial pattern of similarity. 

As expected, \texttt{GPT2MEG} generates data with evoked responses that have much higher correlation with evoked responses from real data, and slightly higher correlation in visual areas compared to other channels, matching the known topography of visual evoked responses. In other models the correlation is low, and spatially better in frontal areas, likely because the evoked responses here are noisier providing an easier fit.

\begin{sloppypar}
Figure~\ref{fig:mean_evoked_corrs} also shows the correlation between the variance (over individual epochs) time-courses of the mean evoked response obtained from the actual data and the evoked responses obtained from data generated by each model. This captures a measure of the ability of the models to represent the trial-to-trial variability found in the real data. Again, \texttt{GPT2MEG} generates data that has the highest correlations with the real data, with higher values in channels in the back of the head, appropriately capturing the topography of response variability. Other models have similar spatial distribution, and notably \texttt{WavenetFullChannel} also produces evoked responses with variance partially matching the real data.
\end{sloppypar}

Finally, a different way to assess task-related activity is to examine the evoked state time-courses from the HMMs fitted on the real and model generated timeseries. Rather than looking at individual channels, this provides an overall view of which HMM state gets activated when, during individual trials. This is computed by simply epoching the state timecourse, and averaging over all trials. We plot these for the real data and each generated timeseries in Figure~\ref{fig:hmm_evokeds}. As expected, the HMM trained on models other than \texttt{GPT2MEG} shows poor evoked state time-courses. \texttt{GPT2MEG} generated data produces states with similar evoked dynamics and variability as the real data. In the next section we show how this generalizes over multiple subjects.

\subsection{Additional ablations and channel-embedding analysis}
\label{ssec:appendix_ablation_embeddings}
\begin{figure}[!t]
  \centering
  \begin{subfigure}{0.33\textwidth}
    \centering
    \includegraphics[width=1.0\linewidth]{forecast_figures/gpt_p80_psd.pdf}
    \caption{Full \texttt{GPT2MEG}}
    \label{fig:gpt_p80_psd}
  \end{subfigure}%
  \begin{subfigure}{0.33\textwidth}
    \centering
    \includegraphics[width=1.0\linewidth]{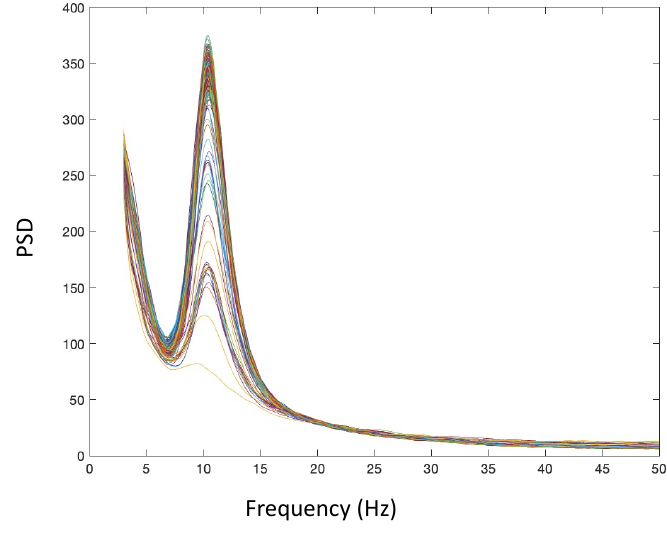}
    \caption{No channel embedding}
    \label{fig:no_channelemb}
  \end{subfigure}%
  \begin{subfigure}{0.33\textwidth}
    \centering
    \includegraphics[width=1.0\linewidth]{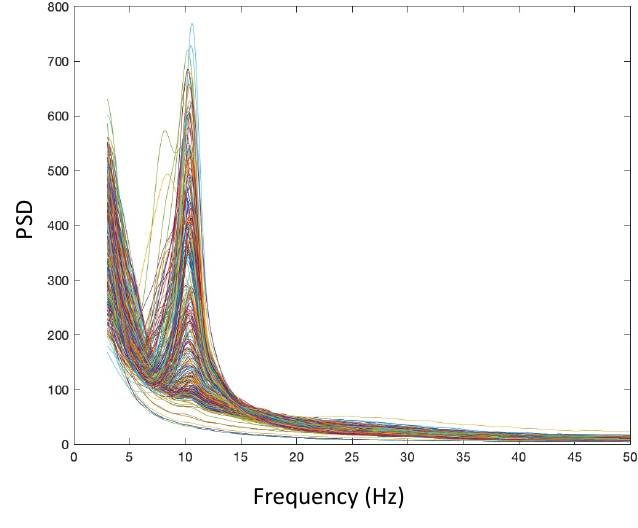}
    \caption{No condition embedding}
    \label{fig:no_condemb}
  \end{subfigure}
  \caption{\textbf{Comparison of generated power spectra with different ablations}: (a) full \texttt{GPT2MEG} model, (b) \texttt{GPT2MEG} with ablated channel embeddings and (c) \texttt{GPT2MEG} with ablated condition embeddings. Shown for a single sample subject. Both channel and condition embeddings are critical for accurate spectral content.}
  \label{fig:embedding_ablation}
  \end{figure}

\begin{figure}[!t]
  \centering
  \begin{subfigure}{0.33\textwidth}
    \centering
    \includegraphics[width=1.0\linewidth]{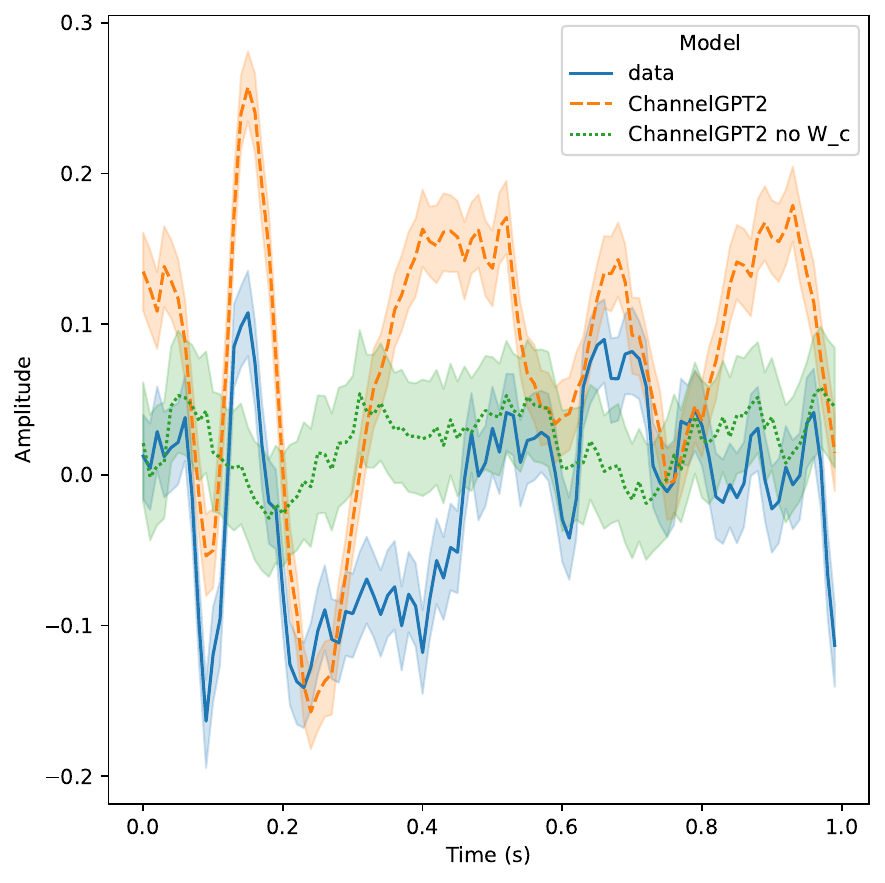}
    \caption{Channel 1 (MEG1011)}
    \label{fig:gpt_ablation_102}
  \end{subfigure}%
  \begin{subfigure}{0.33\textwidth}
    \centering
    \includegraphics[width=1.0\linewidth]{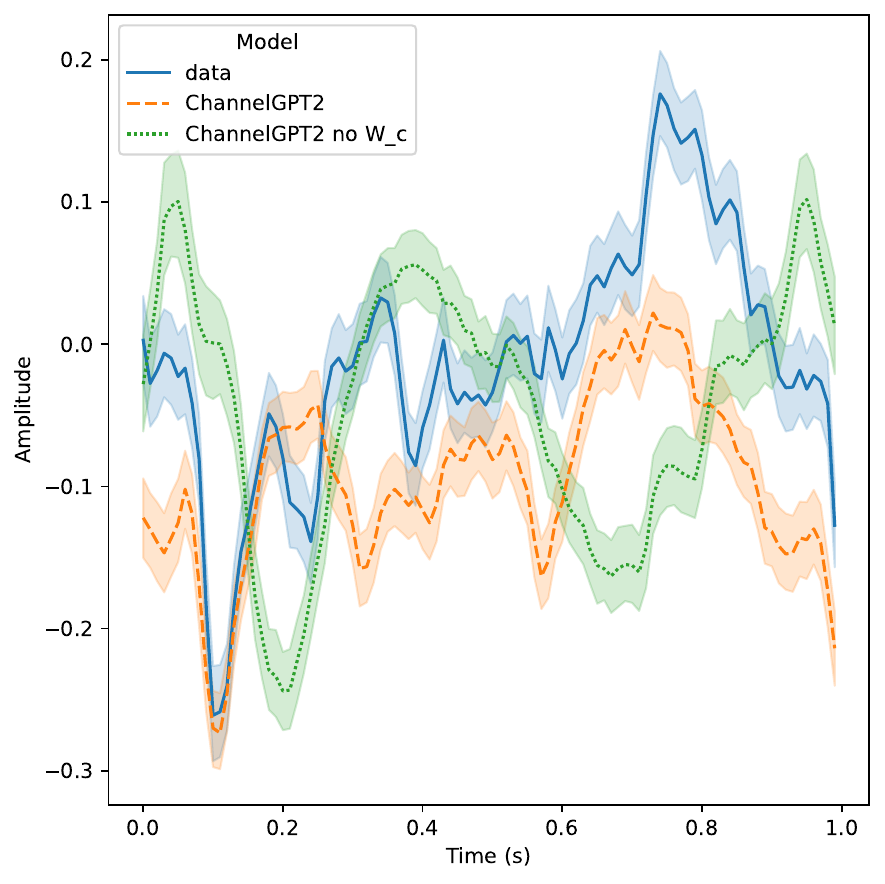}
    \caption{Channel 2 (MEG1813)}
    \label{fig:gpt_ablation_200}
  \end{subfigure}%
  \begin{subfigure}{0.33\textwidth}
    \centering
    \includegraphics[width=1.0\linewidth]{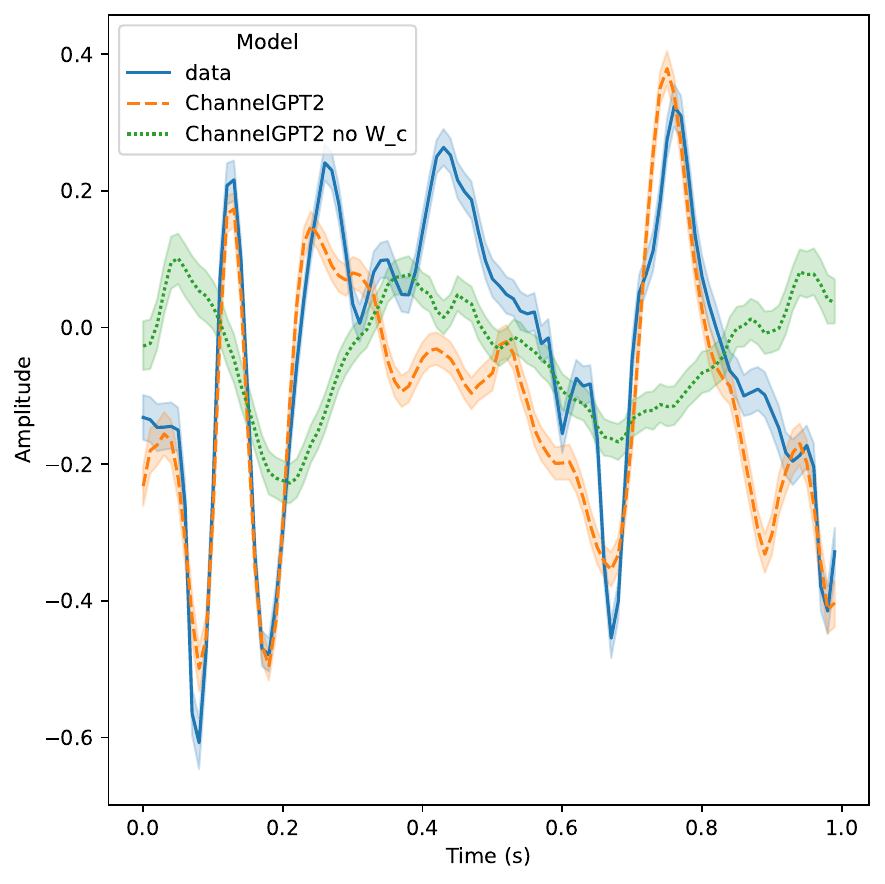}
    \caption{Channel 3 (MEG2332)}
    \label{fig:gpt_ablation_265}
  \end{subfigure}
  \caption{\textbf{Comparison of generated evoked responses with ablated channel embeddings} in the \texttt{GPT2MEG} model, shown across 3 representative channels (a)-(c) and for a single sample subject. Without channel embeddings the model fails to adapt evoked responses to different channels. The stimulus onset is at 0 seconds and the offset is at 0.5 seconds. \texttt{ChannelGPT} refers to \texttt{GPT2MEG}.}
  \label{fig:embedding_ablation_evoked}
  \end{figure}

We also investigated the contributions of the channel and condition embeddings, by training two separate ablated models. As shown in Figure \ref{fig:embedding_ablation}, removing the channel embeddings resulted in very similar PSD across channels in the generated data, indicating that the model relies heavily on these embeddings to adapt generation per channel. The evoked responses in Figure \ref{fig:embedding_ablation_evoked} confirm that without channel embeddings, variability between channels is reduced. Removing the condition embeddings resulted in noisier power spectra of the generated data and no 20 Hz peak.

  \begin{figure}[!t]
  \centering
  \begin{subfigure}{0.33\textwidth}
    \centering
    \includegraphics[width=1.0\linewidth]{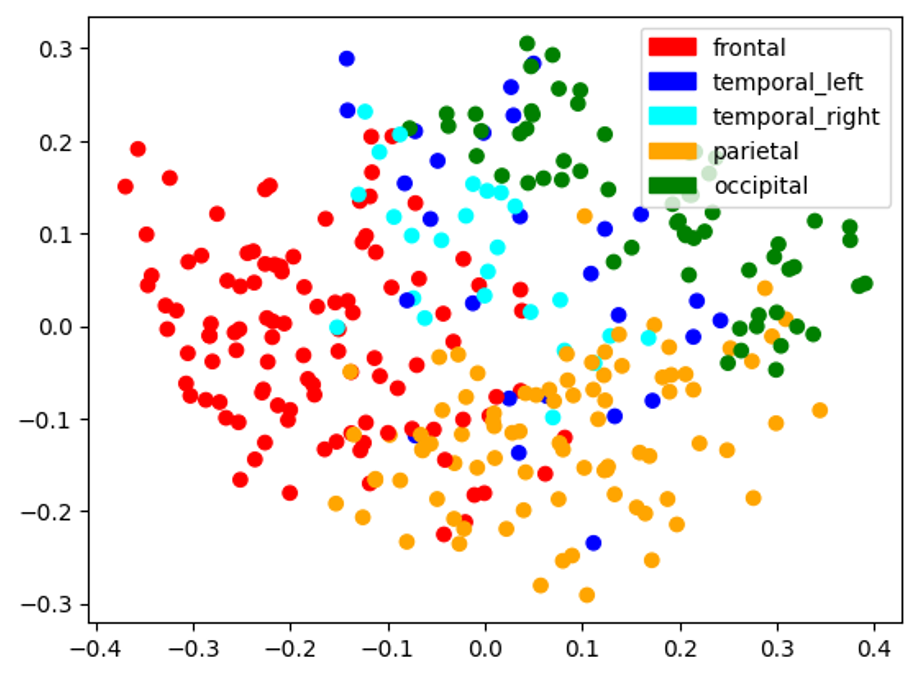}
    \caption{PCA}
    \label{fig:tsne_pca}
  \end{subfigure}%
    \begin{subfigure}{0.33\textwidth}
    \centering
    \includegraphics[width=1.0\linewidth]{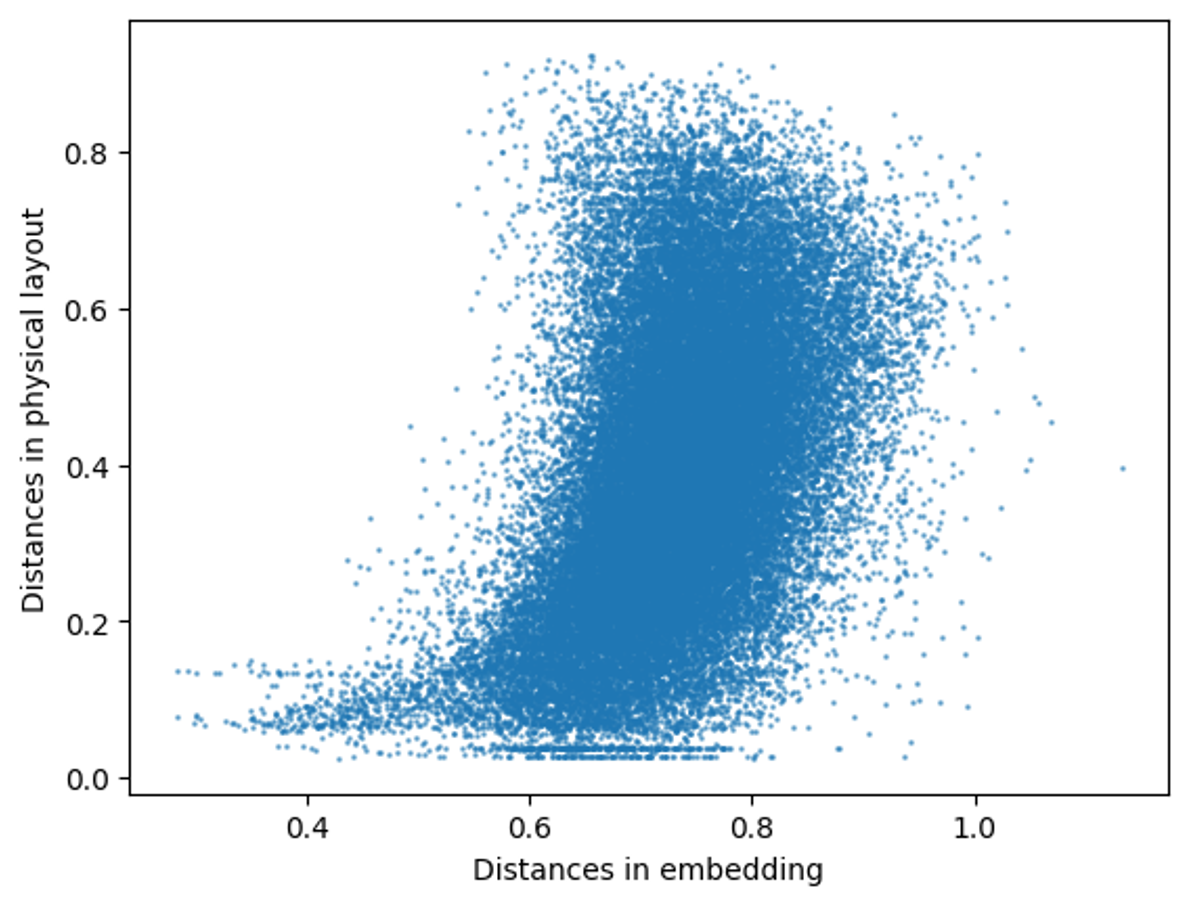}
    \caption{Correlation}
    \label{fig:emb_distances}
  \end{subfigure}%
  \caption{\textbf{Visualisation of channel embeddings.} (a) 2D projection of the channel embeddings from \texttt{GPT2MEG-group} with PCA. Channels are coloured by their location on the scalp grouped into 5 major brain areas. (b) Plotting pairwise Euclidean distances of channels in real, physical space versus embedding space. Sensors that are near to each other in the real sensor montage tend to have more similar embeddings. Each point represents a different pair of channels. Correlation is 0.45.}
  \label{fig:tsne_channelemb}
  \end{figure}

  \begin{figure}[!t]
  \centering
  \begin{subfigure}{0.33\textwidth}
    \centering
    \includegraphics[width=1.0\linewidth]{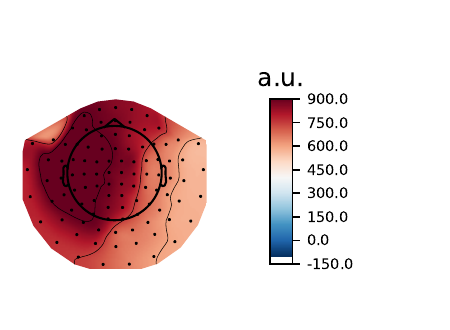}
    \caption{1st component}
    \label{fig:umap1}
  \end{subfigure}%
  \begin{subfigure}{0.33\textwidth}
    \centering
    \includegraphics[width=1.0\linewidth]{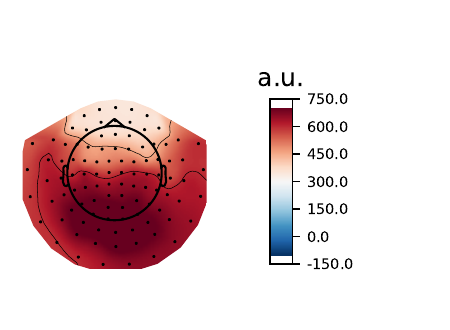}
    \caption{2nd component}
    \label{fig:umap2}
  \end{subfigure}%
   \begin{subfigure}{0.33\textwidth}
    \centering
    \includegraphics[width=1.0\linewidth]{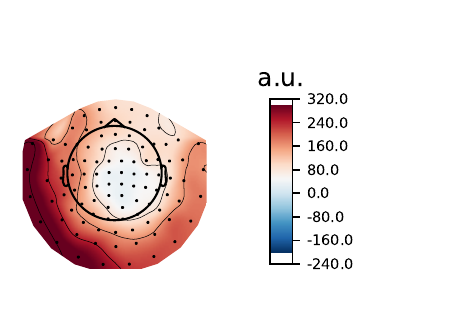}
    \caption{3rd component}
    \label{fig:umap3}
  \end{subfigure}%
  \caption{\textbf{Channel embeddings as sensor plots.} (a) (b) (c) Plotting the first, second, and third components of a UMAP projection of the \texttt{GPT2MEG-group} channel embeddings as sensor plots.}
  \label{fig:umap}
  \end{figure}

Finally, we found that the channel embeddings encode spatial relationships, as sensors that are near to each other in the real sensor montage tend to have more similar embeddings. This is shown through a PCA projection of the embedding space in Figure~\ref{fig:tsne_channelemb}. Correlation between pairwise Euclidean distances of channels in physical space and embedding space was 0.45 (Figure~\ref{fig:emb_distances}). We also plot the first three components of a UMAP projection \citep{McInnes:2018} of the channel embeddings as sensor plots in Figure~\ref{fig:umap}. This clearly shows that the embeddings encode spatial location, with the first and second components encoding the left-right axis and frontal-posterior axis, respectively.

\subsection{Effect of sampling rate on forecasting performance}
\label{ssec:sampling_rate}

\begin{figure}[!t]
\begin{subfigure}{0.49\textwidth}
  \centering
  \includegraphics[width=1.0\linewidth]{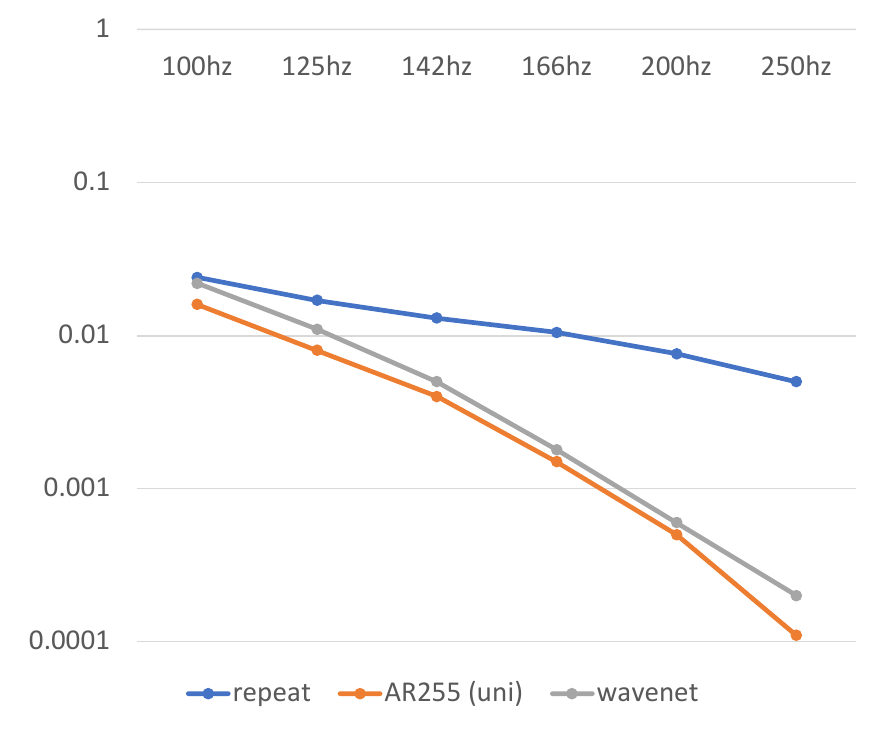}
  \caption{Test MSE}
\end{subfigure}%
\begin{subfigure}{0.49\textwidth}
  \centering
  \includegraphics[width=1.0\linewidth]{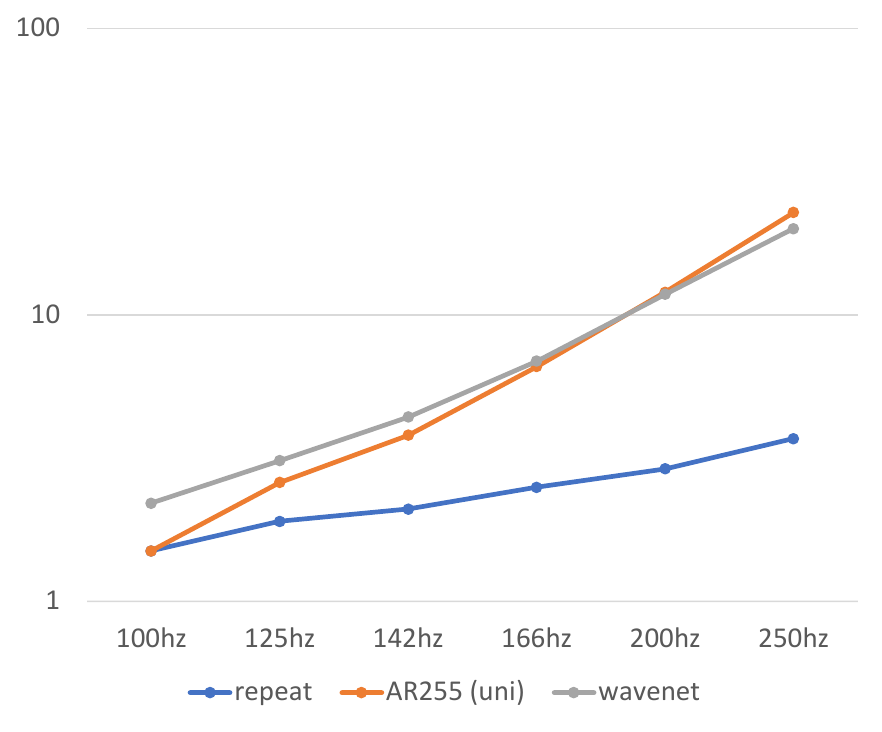}
  \caption{Test accuracy (\%)}
\end{subfigure}
\caption{\textbf{Comparing AR(255) and \texttt{WavenetFullChannelMix} (wavenet) across increasing sampling rates} of the data. \textit{repeat} refers to the repeat baseline. Accuracy is given in percentages.}
\label{fig:sr_comparison}
\end{figure}

We further analysed sampling rate effects on forecasting performance in Figure \ref{fig:sr_comparison}. We trained the AR(255) and \texttt{WavenetFullChannelMix} models on increasing sampling rates of the data from 100 Hz to 250 Hz. The lowpass filter was kept the same at 50 Hz. The receptive fields were kept the same in terms of timesteps, thus they decreased accordingly in terms of actual time in seconds. As expected, both AR and Wavenet models improved markedly with higher sampling rates, as the prediction task became easier when timesteps were closer together. The performance gap between models and the repeating baseline also grew with sampling rate. However, these trends are likely influenced by both the changing prediction interval and filtering artefacts. It is unlikely that such marked improvement would be caused by better modelling of higher-frequency content. Varying the low-pass cut-off with sampling rate reduced performance, suggesting filtering effects dominate. Removal of noise with lower lowpass filters is also a possible explanation. 

\subsection{Generated covariance}
\label{ssec:covariance}

\begin{figure}[!t]
\centering
\begin{subfigure}{0.19\textwidth}
  \centering
  \includegraphics[width=1.0\linewidth]{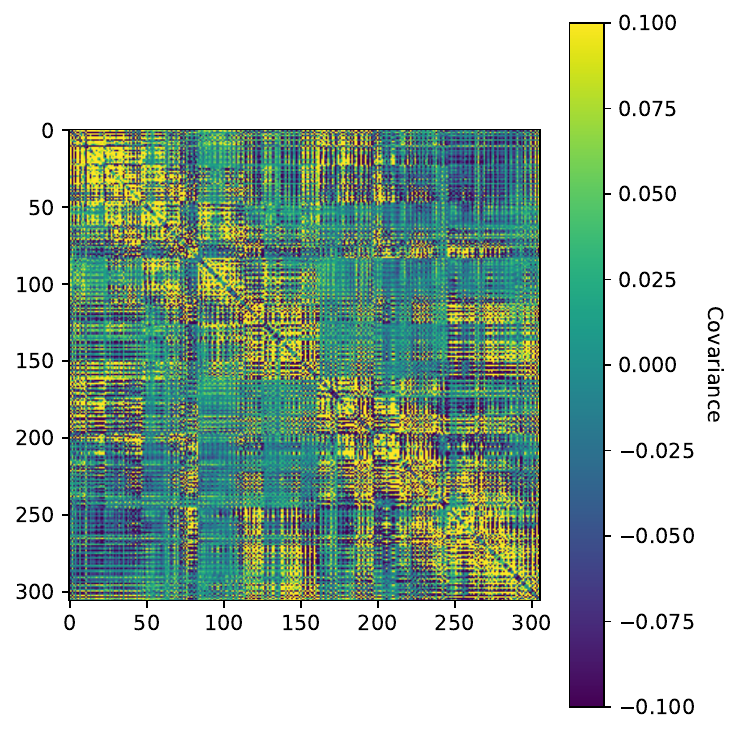}
  \caption{Data}
  \label{fig:data_cov}
\end{subfigure}
\begin{subfigure}{0.19\textwidth}
  \centering
  \includegraphics[width=1.0\linewidth]{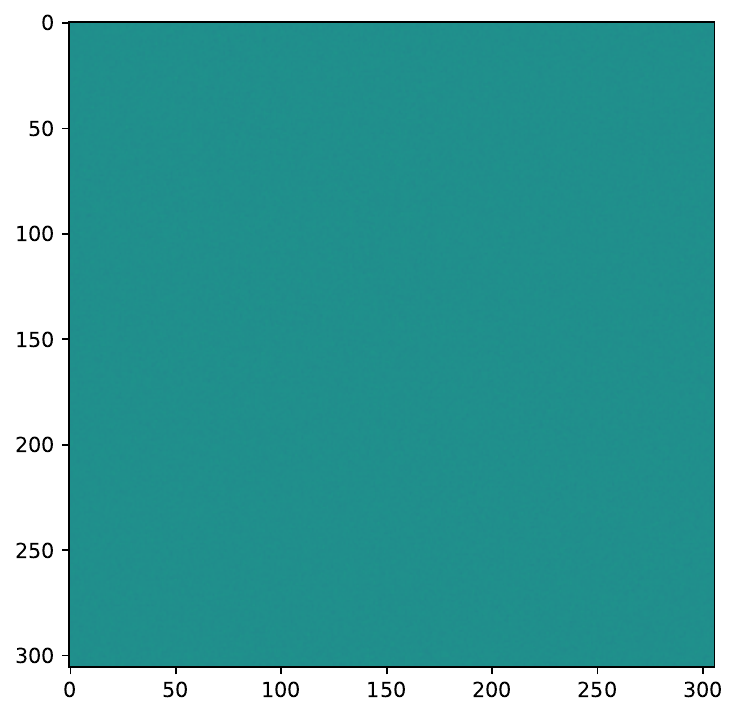}
  \caption{AR(255)}
  \label{fig:ar_cov}
\end{subfigure}
\begin{subfigure}{0.19\textwidth}
  \centering
  \includegraphics[width=1.0\linewidth]{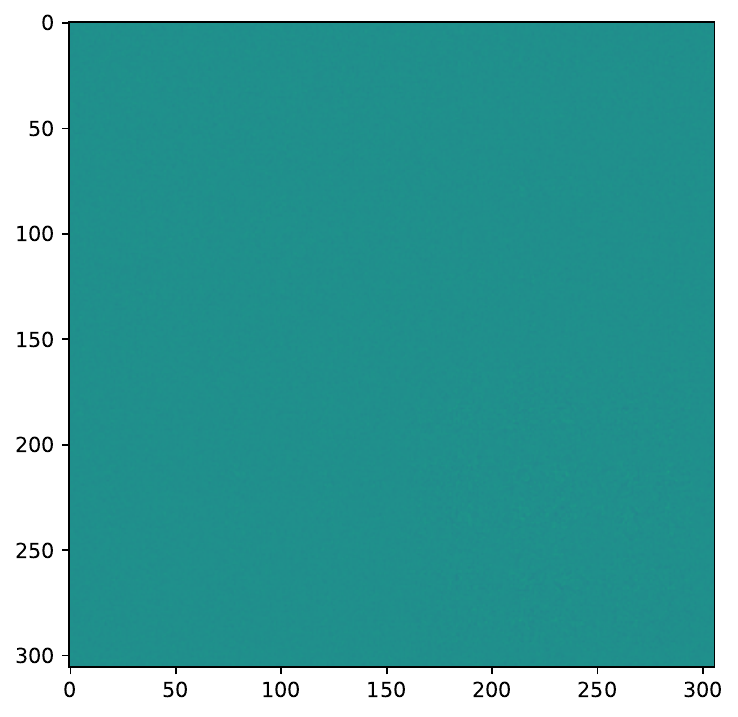}
  \caption{\texttt{WFC}}
  \label{fig:wavenetfullchannel_cov}
\end{subfigure}
\begin{subfigure}{0.19\textwidth}
  \centering
  \includegraphics[width=1.0\linewidth]{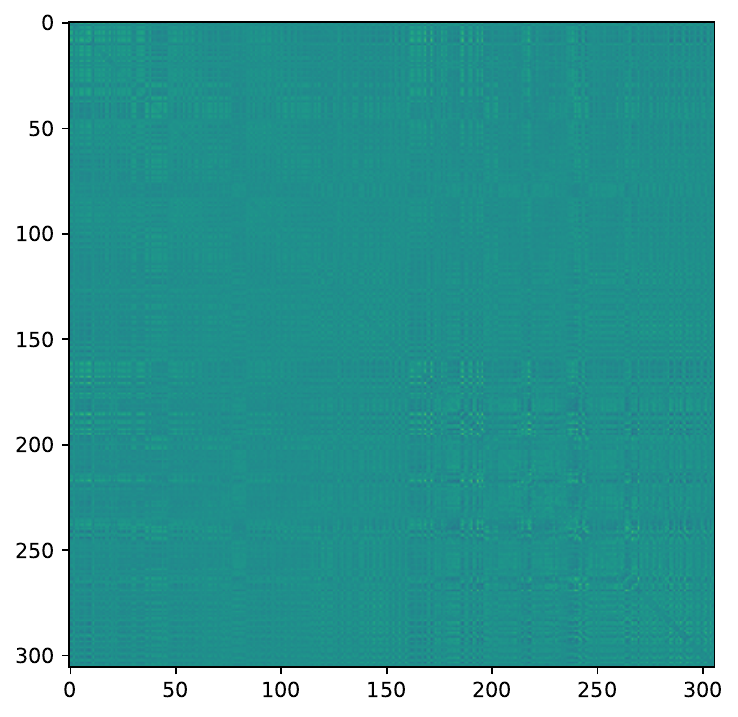}
  \caption{\texttt{GPT2MEG}}
  \label{fig:gpt2_cov}
\end{subfigure}
\begin{subfigure}{0.19\textwidth}
  \centering
  \includegraphics[width=1.0\linewidth]{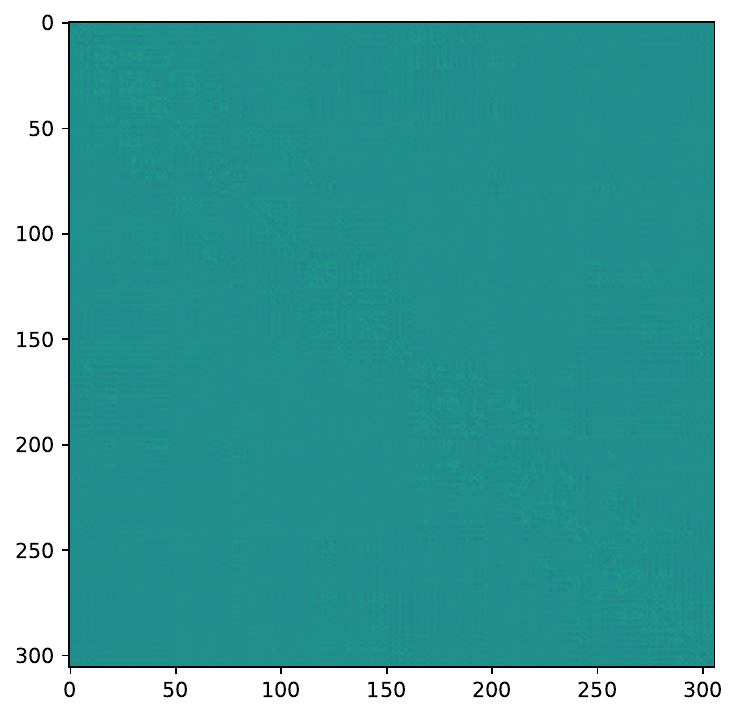}
  \caption{\texttt{WFCM}}
  \label{fig:wavenetfullchannelmix_cov}
\end{subfigure}
\caption{\textbf{Covariance of generated data between channels} (vertical and horizontal axes). All plots have the same scaling as (a).}
\label{fig:generated_cov}
\end{figure}

 As the PSD is a channel-independent measure, we also looked at generated data covariance which captures the interactions between different channels (Figure~\ref{fig:generated_cov}). All models produce data with covariances much closer to 0 than real data. This is perhaps expected for channel-independent models which generate data independently for each channel, but somewhat surprising for \texttt{WavenetFullChannelMix}.

\subsection{\texttt{GPT2MEG-group} evoked analysis}
\label{ssec:group_appendix}

\begin{figure}[!t]
  \centering
  \begin{subfigure}{0.4\textwidth}
    \centering
    \includegraphics[width=1.0\linewidth]{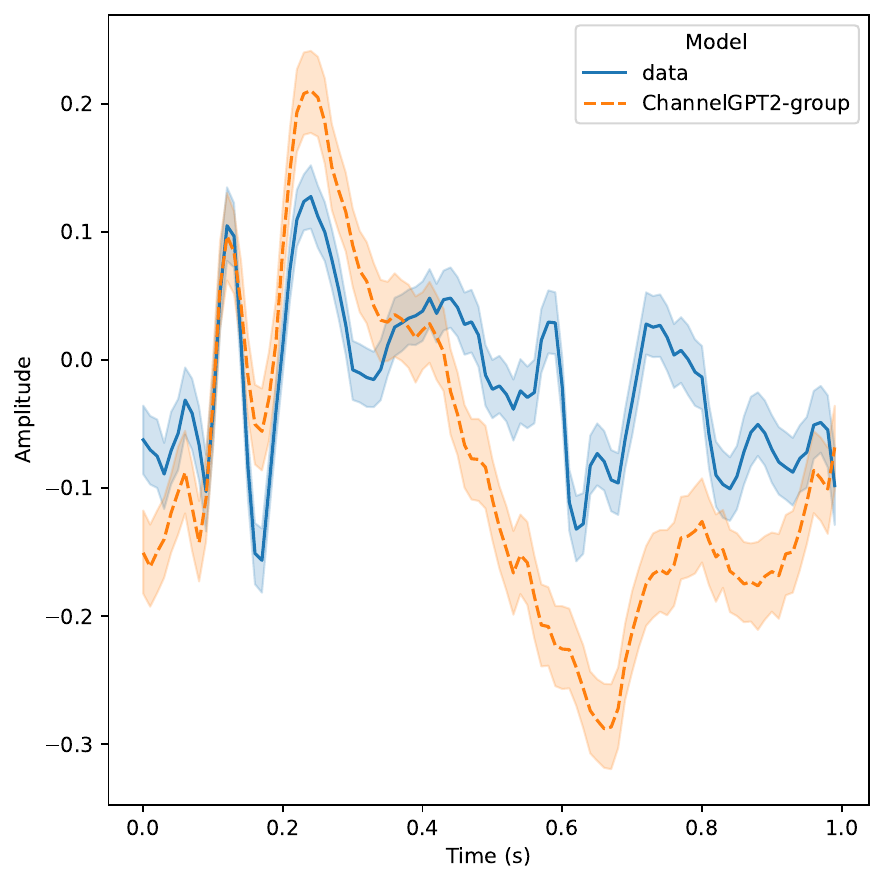}
    \caption{Visual channel (MEG2332)}
    \label{fig:group_data_evoked1}
  \end{subfigure}
  \begin{subfigure}{0.4\textwidth}
    \centering
    \includegraphics[width=1.0\linewidth]{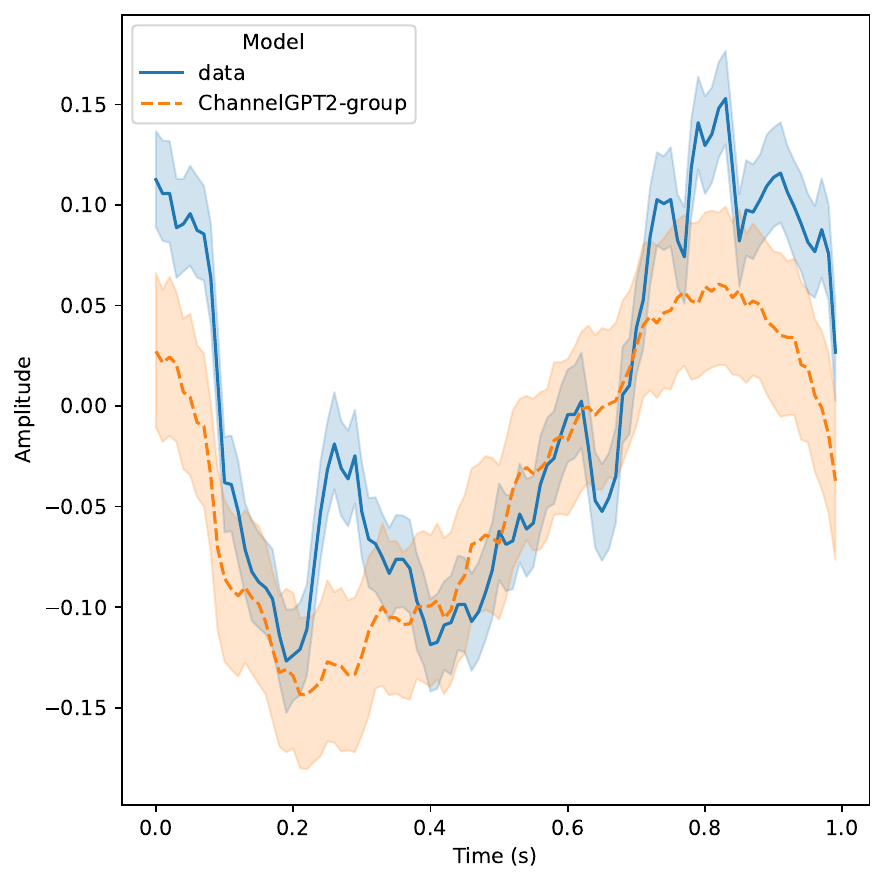}
    \caption{Frontal channel (MEG0111)}
    \label{fig:group_data_evoked2}
  \end{subfigure}
  \caption{\textbf{Comparison of evoked responses averaged across all subjects} in the real MEG data (blue) and across all subjects in the generated data from the \texttt{GPT2MEG-group} model trained on all subjects (orange). The horizontal axis encompasses 1 second, stimulus onset is at 0 seconds and stimulus offset is at 0.5 seconds. Shading indicates 95\% confidence interval of the trial mean. \texttt{ChannelGPT} refers to \texttt{GPT2MEG}.}
  \label{fig:group_evoked_2channels}
  \end{figure}

To test our hypothesis regarding \texttt{GPT2MEG-group} generating more of an average across subjects, we generated data for all subjects (using appropriate subject embeddings) and compared the grand average evoked responses with those extracted from the MEG data of all subjects. Two channels are plotted in Figure~\ref{fig:group_evoked_2channels}. The evoked response averaged over all subjects is much noisier because of the high between-subject variability. However, we can see that indeed \texttt{GPT2MEG-group} can generate this well, perhaps slightly smoother than the real data. Comparing these plots with Figure~\ref{fig:gpt_group_evokeds_visual}, it is also clear that it adapts its generation well to a specific subject compared to the group average. Further evoked activity comparisons based on HMM state timecourses are shown below.

\begin{figure}[!t]
  \centering
  \begin{subfigure}{0.25\textwidth}
    \centering
    \includegraphics[width=1.0\linewidth]{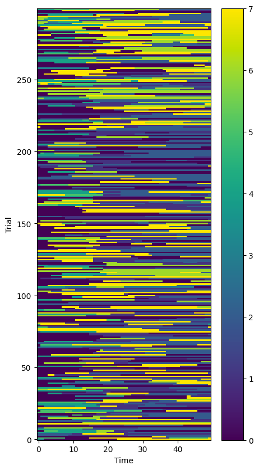}
    \caption{Data}
    \label{fig:data_trialvar}
  \end{subfigure}%
  \begin{subfigure}{0.25\textwidth}
    \centering
    \includegraphics[width=1.0\linewidth]{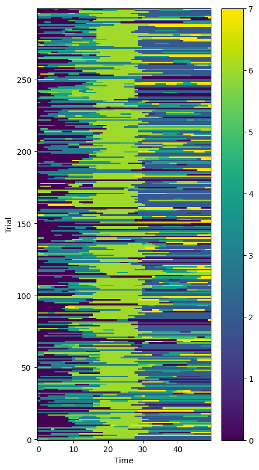}
    \caption{\texttt{GPT2MEG}}
    \label{fig:gpt2_trialvar}
  \end{subfigure}%
  \begin{subfigure}{0.25\textwidth}
    \centering
    \includegraphics[width=1.0\linewidth]{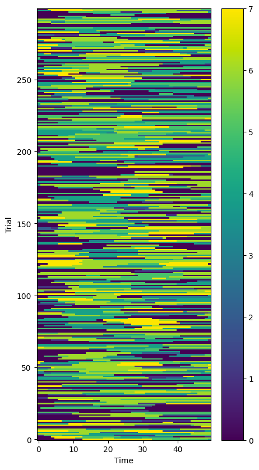}
    \caption{\texttt{GPT2MEG-group}}
    \label{fig:gpt2group_trialvar}
  \end{subfigure}
  \caption{\textbf{Comparison of the trial-level variability in the evoked HMM state timecourses} from an HMM inferred on (a) real MEG data, and on data generated from the (b) \texttt{GPT2MEG} model trained on a single sample subject and (c) \texttt{GPT2MEG-group} trained on all subjects. Different colours represent different states (matched across models). Individual trials however are not matched and we cannot compare the plots at the trial-level, only as an aggregate visualisation of variability across trials.}
  \label{fig:hmm_trialvar}
  \end{figure}

Finally, we examine the variability in state time courses over individual trials. For this we inferred an 8-state HMM on the real data of a single subject. Then, using the HMM observation models held fixed from the real data, we also estimated the state timecourses on both the single-subject \texttt{GPT2MEG} and \texttt{GPT2MEG-group} generated data (with the appropriate subject embedding). Holding the observation models fixed allows us to obtain matched states. We hypothesised that even if the average evoked responses are similar to the real data, GPT2 may not be able to generate trials with variability in the temporal activation of states. Figure~\ref{fig:hmm_trialvar} shows that this is indeed true for the single-subject \texttt{GPT2MEG} generated data. \texttt{GPT2MEG-group}  responses seem to include much higher temporal variability in state activations, though still falling short of the real data. This indicates that the model can capture some trial-to-trial variability through its exposure to multiple subjects, but has difficulty fully matching the complexity of real neural data. More data may be needed to improve this aspect of generation.

\begin{figure}[!t]
  \centering
  \begin{subfigure}{0.49\textwidth}
    \centering
    \includegraphics[width=1.0\linewidth]{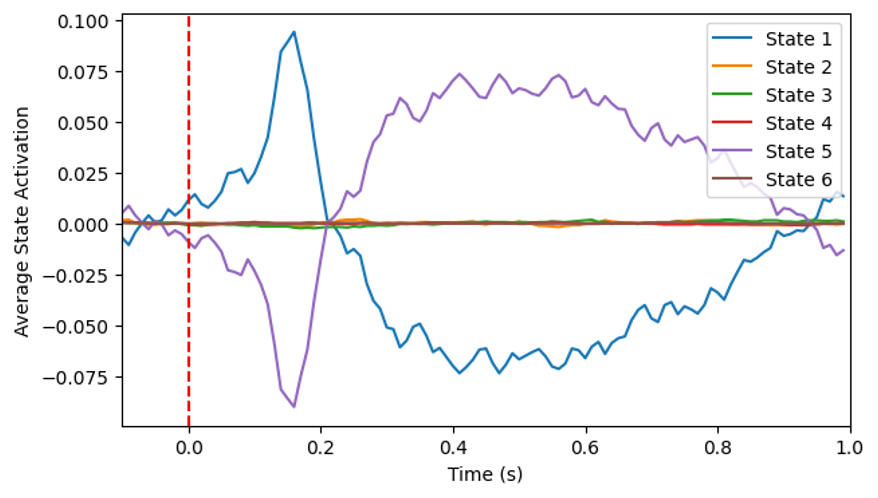}
    \caption{Data}
    \label{fig:data_datahmm}
  \end{subfigure}%
  \begin{subfigure}{0.49\textwidth}
    \centering
    \includegraphics[width=1.0\linewidth]{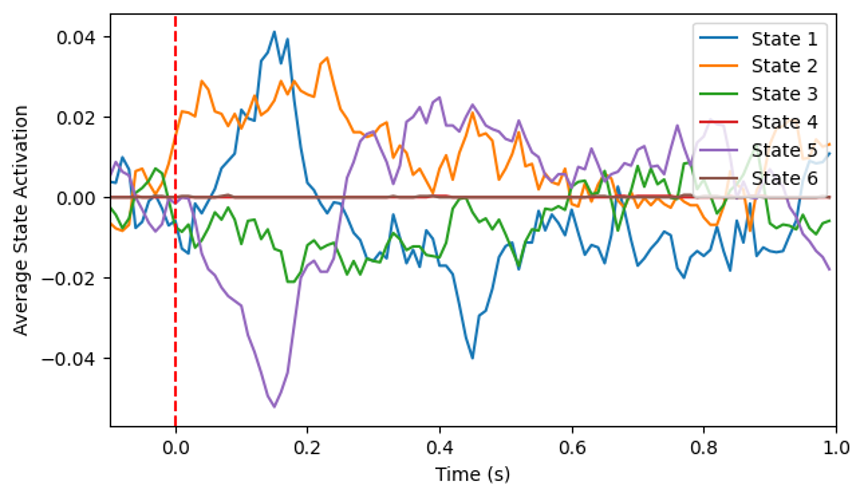}
    \caption{\texttt{GPT2MEG-group}}
    \label{fig:gpt2_datahmm}
  \end{subfigure}
  \caption{\textbf{Comparison of the evoked HMM state timecourses (averaged across all subjects)}; from (a) an HMM inferred on real MEG data, and from (b) data generated from the \texttt{GPT2MEG-group} model trained on all subjects, but where the observation models from the HMM inferred on the real MEG was used. Note that state indices are matched between the two plots, as the same fitted HMM model was used.}
  \label{fig:transfer_hmm_evoked}
  \end{figure}

To further test alignment between group-level evoked responses from a multi-channel perspective, we: 1) inferred an HMM on the real MEG data across all subjects,  2) generated data across all subjects from the trained \texttt{GPT2MEG-group} model, then 3) estimated HMM state timecourses on the generated data, but using the observation models from the HMM inferred in step 1. By holding the observation models fixed we can directly match the evoked HMM state timecourses between the real MEG and generated timeseries. We trained an amplitude-envelope HMM (AE-HMM) with 6 states \citep{quinn2019unpacking}) and show results in Figure~\ref{fig:transfer_hmm_evoked}. Two states that show strong activation during real task data show similar temporal signatures and amplitude changes in the generated data, albeit slightly noisier. In the generated data, there are two additional states which seem to get activated during the trial. This indicates that while \texttt{GPT2MEG-group} can capture some of the state-level dynamics, there is room for improvement.

\subsection{\texttt{GPT2MEG} adapts generation to different trial lengths}
\label{ssec:diff_trial_length}
\begin{figure}[!t]
    \centering
    \includegraphics[width=0.5\textwidth]{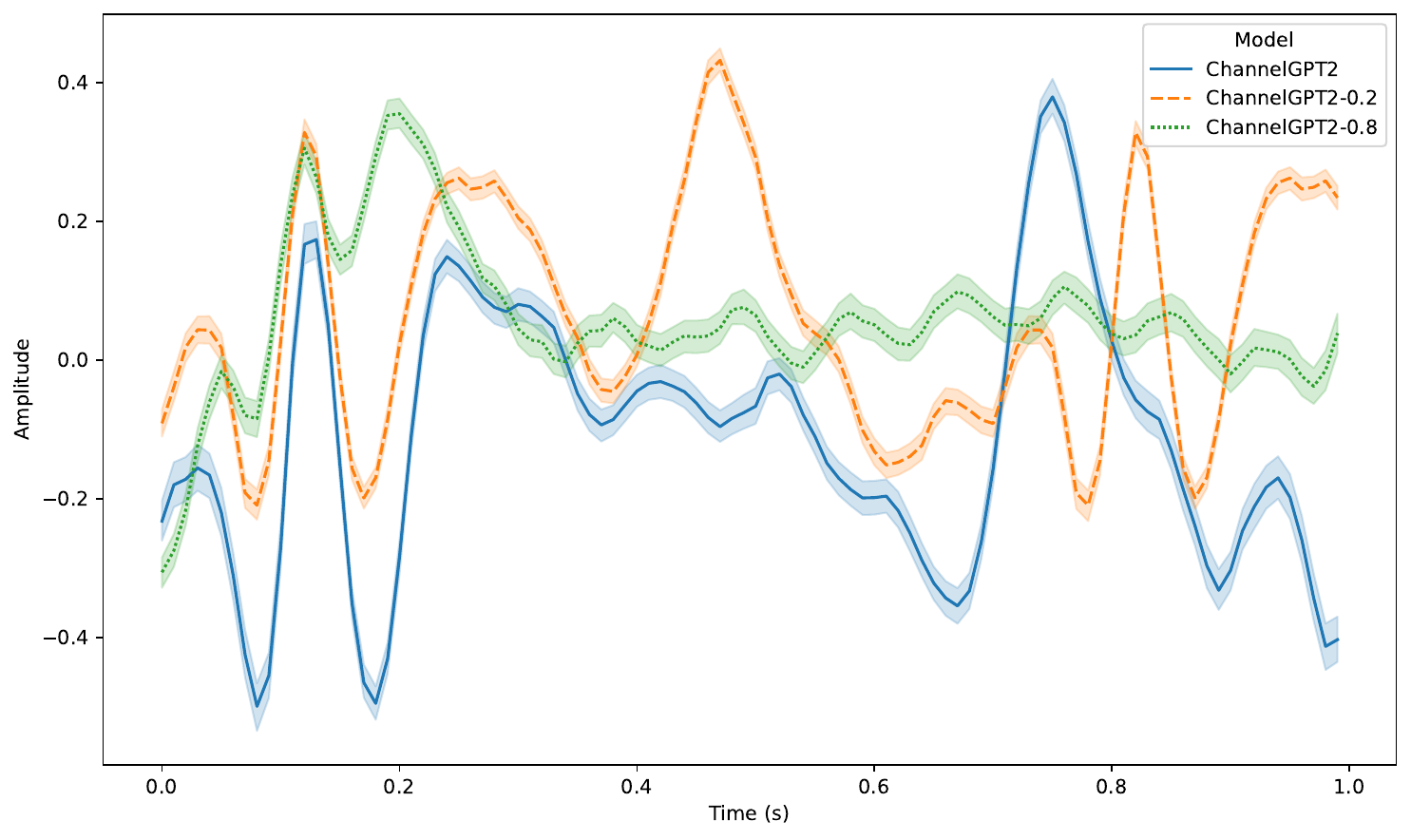}
    \caption{\textbf{Evoked responses generated by GPT2MEG} for a single sample subject, for trials of 0.2 s (orange), 0.5 s (blue), and 0.8 s (green). While the model was trained only on data containing trials of 0.5 s, it shows an ability to adapt appropriately to the different durations. The plotted channel is in the visual area (MEG2332). \texttt{ChannelGPT} refers to \texttt{GPT2MEG}.}
    \label{fig:evoked_random_timing}
\end{figure}

We evaluated the model’s ability to adapt to different trial durations. The results reported thus far are for a \texttt{GPT2MEG} trained on trials lasting 0.5 seconds. We generated data using the same fitted \texttt{GPT2MEG} model but with trial durations of 0.2 s and 0.8 s. As shown in Figure \ref{fig:evoked_random_timing}, \texttt{GPT2MEG} accurately adapted to the shorter and longer trials. The evoked responses matched the expected time-courses, with appropriate truncation or lack of second peaks due to stimulus offset. This demonstrates the model’s ability to generalise to varied trial durations despite being trained on a fixed duration.

\end{document}